Thesis for Master's Degree

# Mem-elements based Neuromorphic Hardware for Neural Network Application

Ankur Singh

School of Electrical Engineering and Computer Science

Gwangju Institute of Science and Technology (GIST)

2024

# Mem-elements based Neuromorphic Hardware for Neural Networks Application

# 신경망 애플리케이션을 위한 메모리 요소 기반 뉴로모픽 하드웨어

# Mem-elements based Neuromorphic Hardware for Neural Networks Application

Advisor: Professor Byung-Geun Lee

By

Ankur Singh

School of Electrical Engineering and Computer Science

Gwangju Institute of Science and Technology

A thesis submitted to the faculty of the Gwangju Institute of Science and Technology in partial fulfilment of the requirements for the degree of Master of Science in the School of Electrical Engineering and Computer Science.

Gwangju, Republic of Korea

2023. 12. 14.

Approved by

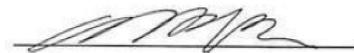

Professor Byung-Geun Lee

Committee Chair

# Mem-elements based Neuromorphic Hardware for Neural Networks Application

Ankur Singh

Accepted in partial fulfillment of the requirement

for the degree of Master of Science

December 14, 2023

Committee Chair 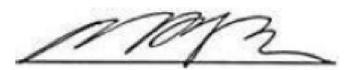
Prof. Byung-Geun Lee

Committee Member 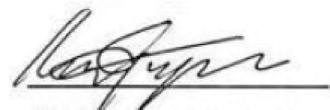
Prof. Moongu Jeon

Committee Member 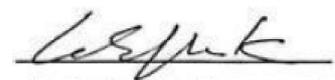
Prof. Chang-Soo Park

## Abstract

The thesis investigates the utilization of memristive and memcapacitive crossbar arrays in low-power machine learning accelerators, offering a comprehensive co-design framework for deep neural networks (DNN). The model, implemented through a hybrid Python and PyTorch approach, accounts for various non-idealities, achieving exceptional training accuracies of 90.02% and 91.03% for the CIFAR-10 dataset with memristive and memcapacitive crossbar arrays on an 8-layer VGG network. Additionally, the thesis introduces a novel approach to emulate meminductor devices using Operational Transconductance Amplifiers (OTA) and capacitors, showcasing adjustable behavior. Transistor-level simulations in 180 nm CMOS technology, operating at 60 MHz, demonstrate the proposed meminductor emulator's viability with a power consumption of 0.337 mW. The design is further validated in neuromorphic circuits and CNN accelerators, achieving training and testing accuracies of 91.04% and 88.82%, respectively.

Notably, the exclusive use of MOS transistors ensures the feasibility of monolithic IC fabrication. This research significantly contributes to the exploration of advanced hardware solutions for efficient and high-performance machine-learning applications.

*Dedicated to my family*

# Contents

# List of Figs

# List of Tables

# Chapter 1

# Introduction

## 1.1 Motivation

In the contemporary landscape, the proliferation of sophisticated artificial intelligence (AI) algorithms across applications such as smart devices, self-driving cars, and surveillance systems has precipitated a surge in demand for hardware platforms capable of efficiently executing artificial neural network (ANN) algorithms. Various hardware platforms, including graphics processing units (GPUs), field-programmable gate arrays (FPGAs), and application-specific integrated circuits (ASICs), have been employed to implement AI hardware. While GPU/FPGA hardware exhibits high compatibility and reconfigurability across different applications, its substantial power consumption limits its utility in mobile and edge computing. In contrast, ASIC hardware platforms offer lower power consumption but are not immune to the von Neumann bottleneck, particularly when handling the substantial data movement between memory blocks and processors as the depth of ANN layers increases. The exponential growth of data and the escalating complexity of neural network models underscore the constraints of traditional computing architectures in terms of power efficiency, speed, and scalability [1-2]. In response to these challenges, neuromorphic computing has emerged as a promising alternative, drawing inspiration from the brain's architecture to design more efficient and brain-like computing systems. A significant breakthrough in neuromorphic computing involves the application of mem-elements, seamlessly integrating memory and processing functions.

In the realm of artificial intelligence (AI), the relentless advancement of algorithms has been met with a growing demand for hardware platforms capable of efficiently executing complex neural network models. As traditional computing architectures face challenges related to power efficiency, speed, and scalability, the emergence of neuromorphic computing has garnered attention for its potential to revolutionize the way we approach AI hardware [3]. Within this paradigm, the integration of mem-elements, including memristors, memcapacitors, and the innovative introduction of meminductors, as shown in Fig. 1.1 using complementary metal-oxide-semiconductor (CMOS) technology, presents a promising avenue for enhancing the capabilities of neural networks. The conventional von Neumann architecture, characterized by the separation of memory and processing units, gives rise to energy and time inefficiencies during the movement of vast amounts of data between these components. The utilization of mem-elements, such as memristors and memcapacitors, addresses this bottleneck by enabling localized processing within memory [4]. This departure from traditional architectures aligns with neural computations' distributed and parallel nature, offering a novel and efficient approach to hardware design.

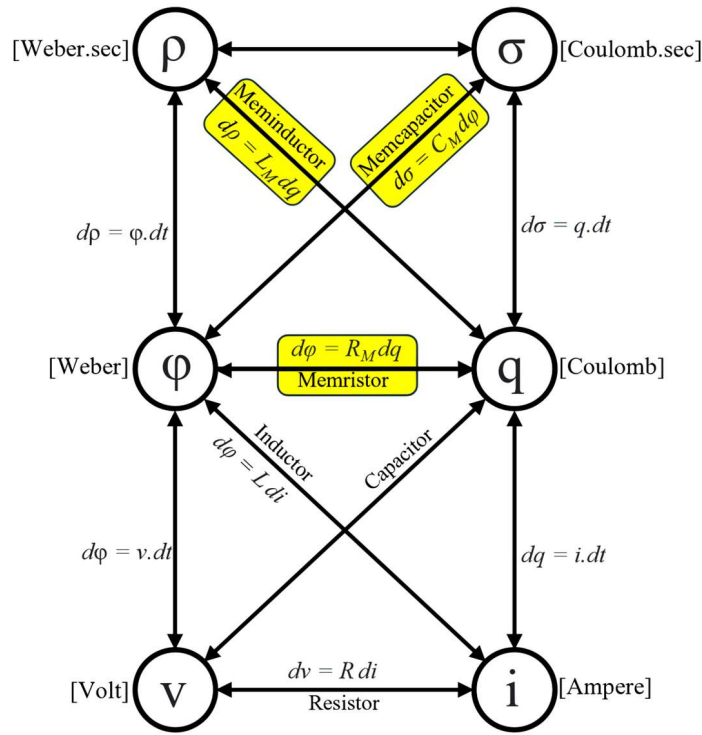

**Fig. 1.1 Memory and non-memory components relationship**

The motivation behind exploring mem-elements-based neuromorphic hardware for neural network applications is rooted in the need for energy-efficient and high-performance computing solutions. The challenges posed by modern AI applications, particularly those involving deep learning and massive data processing, necessitate innovative hardware solutions. The incorporation of memristors and memcapacitors, alongside the novel application of meminductors, has the potential to fundamentally transform the landscape of AI hardware, overcoming the limitations imposed by traditional computing architectures as shown in Fig. 1.2. The integration of real hardware data from memristor and memcapacitor devices into Python-based neural network implementations adds a practical and tangible dimension to this research [5]. By leveraging actual hardware performance data, the study validates the theoretical concepts and provides a bridge between simulation and real-world application, enhancing the robustness and applicability of the proposed neuromorphic hardware.

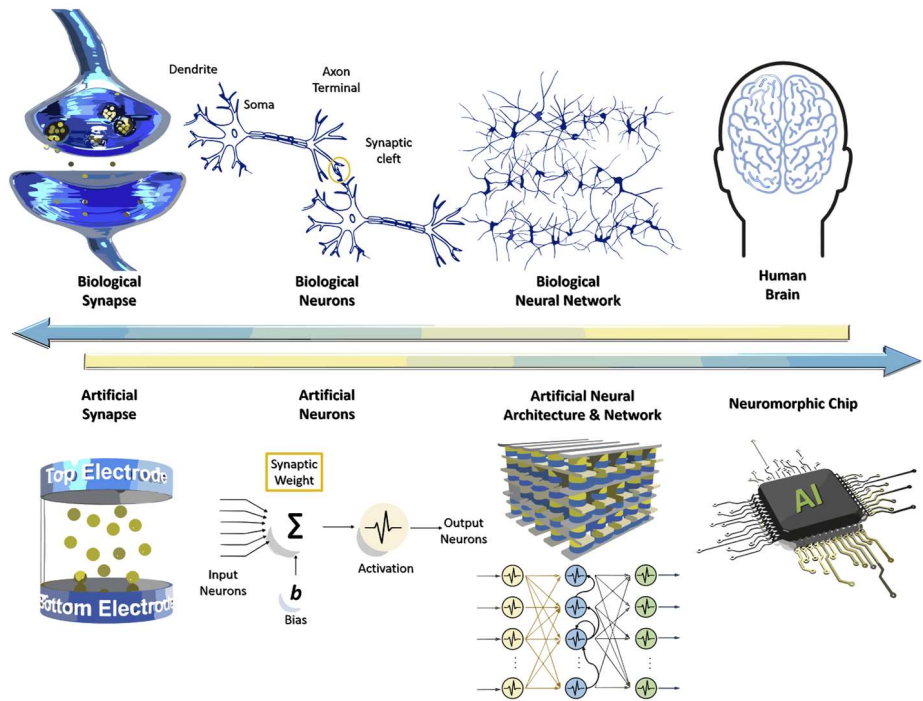

**Fig. 1.2 Concept of artificial neural system in neuromorphic devices**

Furthermore, the design and implementation of meminductors using CMOS technology represent a forward-thinking approach to expanding mem-elements' repertoire. This innovation showcases the versatility of neuromorphic hardware design and introduces a new element that contributes to neural networks' overall efficiency and adaptability. The significance of this research extends to edge computing, where power efficiency is paramount. Mem-elements-based neuromorphic hardware has the potential to redefine the capabilities of edge devices, enabling them to perform complex AI tasks locally without overreliance on cloud resources. This aligns with the contemporary trend towards decentralized computing and contributes to the development of more autonomous and intelligent edge devices.

**Chapter 1.2 Thesis Organization**

This thesis is structured as follows: Chapter 2 provides an introduction to the fundamentals of mem-elements. In Chapter 3, we delve into the discussion of memristor and memcapacitor vector-matrix multiplication for on-chip training, considering non-idealities. Chapter 4 covers the design of CMOS meminductor and introduces neuromorphic hardware applications based on meminductor. Moving on to Chapter 5, we describe the simulation of all the mem-elements with various analyses. Finally, Chapter 6 offers concluding remarks for the thesis.

# Chapter 2

# Basics of Mem-elements

In the relentless pursuit of advancing electronic technologies, the realm of mem-elements stands as a testament to innovation and transformative potential. Coined from "memory-enhanced elements," mem-elements represent a groundbreaking class of electronic components that introduce a novel paradigm by seamlessly integrating memory functionalities into traditional passive circuit elements. This burgeoning field challenges conventional paradigms and reshapes the landscape of electronic engineering, promising a paradigm shift in the design and functionality of electronic systems. At its core, mem-elements encompass a diverse class of components, with each member contributing uniquely to the integration of memory into electronic circuits. The triad of memristors, memcapacitors, and meminductors serves as the vanguard of this technological revolution, each bringing its distinct attributes to the forefront of innovation.

Complementing the memristor, memcapacitors introduce dynamic memory capabilities to traditional capacitors. These components dynamically adjust their capacitance based on the historical patterns of applied voltage, paving the way for adaptive circuits, signal processing advancements, and energy-efficient electronic systems. The marriage of memory and capacitance represents a significant departure from traditional electronic components, opening up new possibilities for responsive and versatile circuits. The meminductor, a more recent addition to the mem-elements family, extends the concept of memory-enhanced elements by fusing memory characteristics with inductive components [6]. This innovative element introduces memory-enhanced inductance, offering unique prospects for unconventional computing paradigms and signal-processing applications.

## 2.1 Memristor

Memristor devices, short for "memory resistor," represent a revolutionary class of electronic components that have redefined our understanding of resistance and memory within the framework of electronic circuits. Coined by Professor Leon Chua in 1971 [7], the memristor is the fourth fundamental circuit element, alongside resistors, capacitors, and inductors, challenging and expanding the traditional boundaries of electronic engineering. At its essence, a memristor is characterized by its ability to store and recall a specific resistance value based on the history of the applied voltage. The memristor symbol and the hysteresis curve are shown in Fig. 2.1(a) and 2.2. This unique property makes memristors pivotal in the development of neuromorphic computing, a field inspired by the architecture and functionality of the human brain. In biological systems, synapses exhibit plasticity, adjusting their strength based on prior electrical activity. Memristors emulate this synaptic plasticity, enabling the creation of artificial neural networks that can adapt and learn from input patterns.

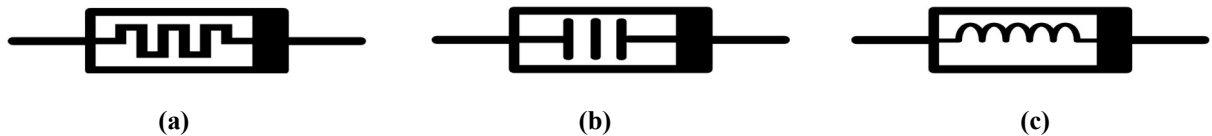

(a)            (b)            (c)

**Fig. 2.1 (a) Memristor (b) memcapacitor (c) meminductor symbol**

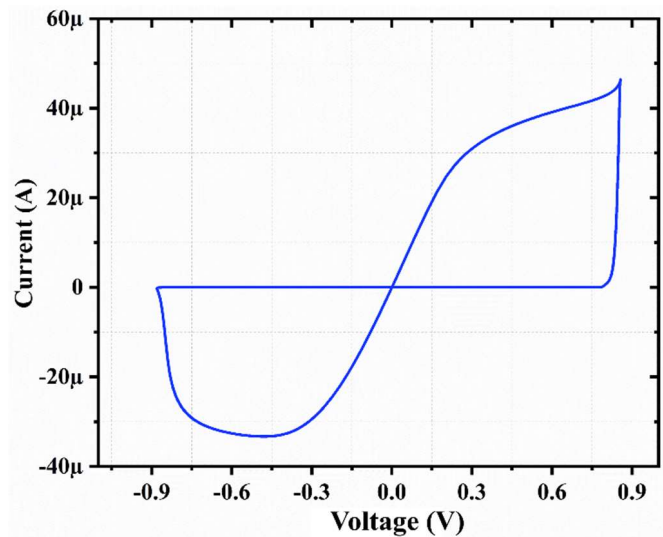

**Fig. 2.2 Current-voltage pinched hysteresis curve of memristor**

The fundamental operation of a memristor involves changes in resistance, typically induced by the flow of charge carriers through a thin insulating layer. This change in resistance can be controlled and manipulated, creating non-volatile memory devices that retain their resistance state even when power is removed. This non-volatility is a key advantage compared to traditional volatile memory technologies, offering potential data storage and processing applications. Moreover, the nanoscale dimensions of memristor devices make them attractive for future generations of electronic devices. Their compact size, low power requirements, and potential for integration with complementary metal-oxide-semiconductor (CMOS) technology contribute to their appeal in the development of next-generation computing architectures. Applications of memristors extend beyond computing and memory.

They hold promise in analogue signal processing, brain-inspired computing, and even in the emulation of complex biological systems for medical research. The ability of memristors to dynamically adjust their resistance based on prior electrical activity positions them as versatile components with the potential to revolutionize various facets of electronic engineering.

**2.2 Memcapacitor**

Memcapacitor devices, a portmanteau of "memory" and "capacitor," constitute a novel class of electronic components that extend the capabilities of traditional capacitors by introducing memory functionalities. This innovative concept represents a significant departure from conventional passive circuit elements and has the potential to reshape the landscape of electronic engineering. Memcapacitors are distinguished by their ability to dynamically alter their capacitance based on the historical patterns of applied voltage, thereby introducing a memory component into capacitance behaviour and shown in are shown in Fig. 2.1(b) and 2.3. In essence, a memcapacitor exhibits a capacitance that is not solely determined by the instantaneous voltage but is influenced by the electrical history of the device. This unique property opens up avenues for adaptive electronic systems, responsive circuits, and innovative signal-processing applications. The fundamental operation of a memcapacitor involves the modulation of its dielectric properties, typically achieved through the incorporation of materials with variable permittivity or through other mechanisms that enable dynamic changes in capacitance.

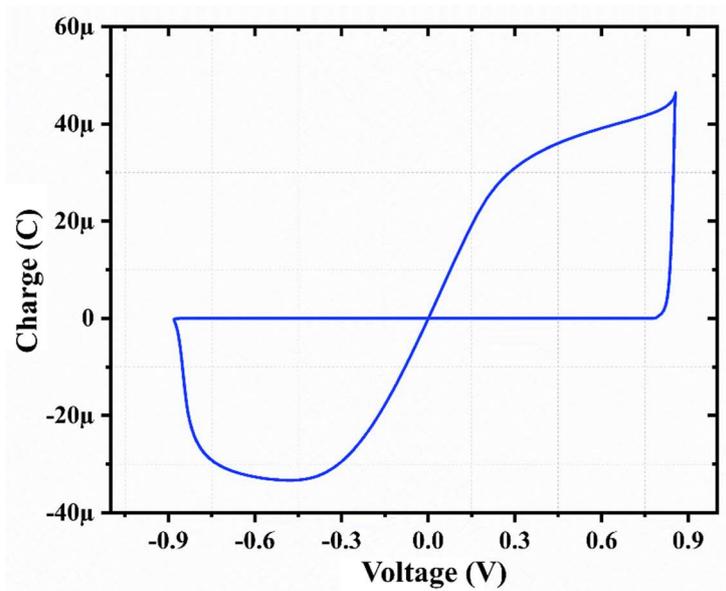

**Fig. 2.3 Current-voltage pinched hysteresis curve of memristor.**

One of the key advantages of memcapacitor devices lies in their potential applications in adaptive circuits. By virtue of their memory-enhanced capacitance, these devices can be employed in signal-processing applications where the response of a circuit is influenced by past input patterns. This capability is particularly advantageous in fields such as pattern recognition, where the ability to adapt to changing input conditions is crucial. Moreover, memcapacitors have potential implications for energy-efficient electronic systems. Their adaptive nature allows for the optimization of energy consumption by dynamically adjusting capacitance in response to varying operational requirements. This property makes them appealing for applications in low-power electronics and energy harvesting, contributing to the development of sustainable electronic technologies.

**2.3 Meminductor**

Meminductor devices, a fusion of "memory" and "inductor," represent an exciting frontier in electronic engineering, introducing a novel class of components that seamlessly integrate memory functionalities with inductive properties. This innovative concept expands upon traditional passive circuit elements and opens up new possibilities for the development of memory-enhanced inductive components and shown in Fig. 2.1(c) and 2.4. At its core, a meminductor exhibits both inductance and memory characteristics, distinguishing it from conventional inductors. Inductance, a property that resists changes in current flow, is augmented by the ability of the meminductor to store and recall specific inductance values based on the historical patterns of applied voltage. This dual functionality introduces a dynamic and adaptable element into electronic circuits, offering the potential for innovative applications in signal processing, unconventional computing paradigms, and beyond. The operation of a meminductor involves the modulation of inductance values through mechanisms such as changes in magnetic permeability or alterations in the geometric configuration of the device. This dynamic control over inductance, coupled with the ability to retain specific inductance states, positions meminductors as versatile components with implications for diverse electronic applications.

One of the key advantages of meminductor devices lies in their potential to contribute to the development of adaptive and memory-enhanced electronic systems. Their ability to store and recall specific inductance values enables the creation of circuits that can adapt to varying input conditions, making them valuable in applications where responsiveness and memory are critical. Meminductors also hold promise in unconventional computing paradigms, contributing to the exploration of alternative approaches to information processing. The integration of memory with inductive elements introduces a unique dynamic to circuit behaviour, fostering the development of systems that go beyond traditional binary logic. As with other mem-elements, the nanoscale dimensions of meminductor devices make them attractive for future electronic devices. Their compact size and potential integration with complementary metal-oxide-semiconductor (CMOS) technology contribute to their appeal for next-generation electronic architectures.

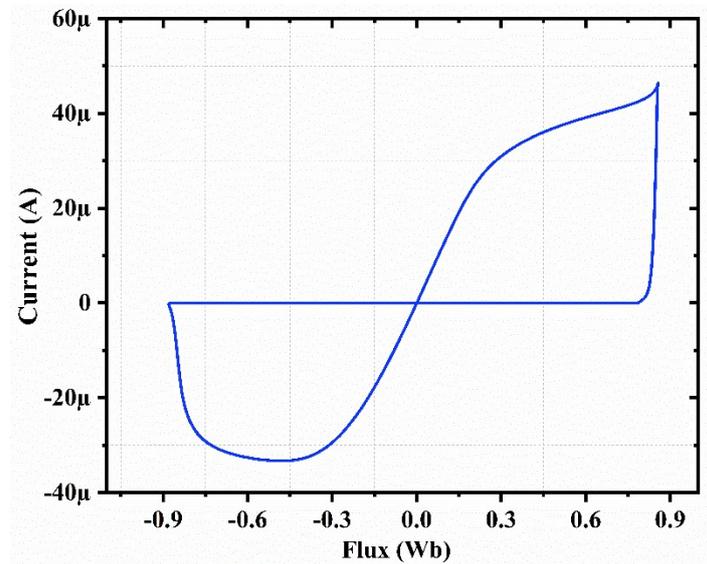

**Fig. 2.4 Current-voltage pinched hysteresis curve of memristor**

# Chapter 3

# Proposed Memristor and Memcapacitor for On-Chip Training

In this chapter, we extended the NeuroSim framework to support the evaluation of on-chip training performance in compute-in-memory (IMC) accelerators [8]. The framework is implemented using a hybrid approach with Python and PyTorch CUDA together, incorporating an accurate VMM computation core [9]. The performance of our proposed framework was validated by implementing a simplified 8-layer VGG network with a 128 × 128 $TiO_X$-based memristive and a Silicon (Si)-based memcapacitive crossbar array, both derived from hardware data. The results demonstrated an impressive 90.02% and 91.03% inference accuracy with memristive and memcapacitive crossbar array on the CIFAR10 dataset. Our proposed method offers a potential solution for holistically evaluating neural network performance and highlights the feasibility of utilizing TiOx-based memristive and Si-based memcapacitive crossbars for DNN applications.

## 3.1 Proposed On-Chip Training Framework

### 3.1.1 In-memory Computing Mapping Algorithm

In this approach, the weights of each kernel are intelligently divided into sub-matrices based on their spatial locations, resulting in K×K sub-matrices with a size of D×N. This division leads to a total weight matrix size which is kernel of K×K×D×N, as shown in Fig. 3.1. Simultaneously, the input data assigned to various spatial locations within each kernel is routed to the corresponding sub-matrices. Through parallel computation, partial sums are derived from these sub-matrices. These partial sums are then efficiently aggregated using an adder tree. By doing so, a processing element (PE) is defined as a group of sub-arrays equipped with essential input and output buffers, along with accumulation modules. The kernels are divided into multiple PEs based on their spatial locations, allowing the input data to be assigned accordingly. This strategic division of kernels and input data enables the reusability of input data among PEs, removing the need to revisit upper-level buffers. Consequently, a direct transfer of input data between PEs is facilitated, streamlining the processing flow, and optimizing computational efficiency.

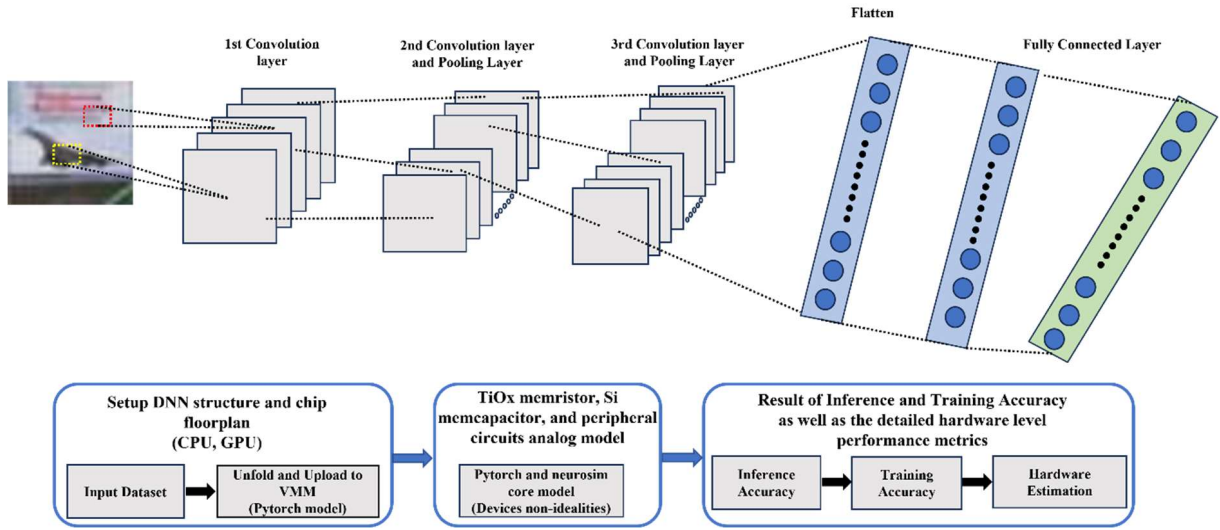

**Fig. 3.1 Concept of artificial neural system in neuromorphic devices**

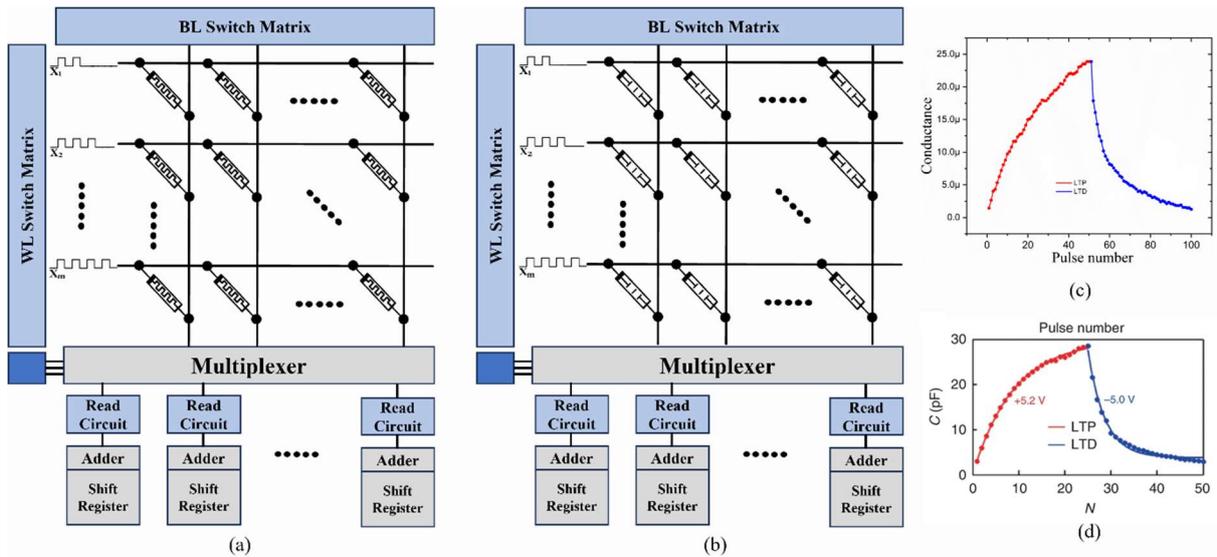

**Fig. 3.2 Concept of artificial neural system in neuromorphic devices**

The chip hierarchy is organized into several tiles, each of which houses processing elements (PEs) along with synaptic sub-arrays, accumulation modules, and output buffers. The transfer of inputs/activations from one memory array to another is facilitated by interconnects within each tile. In terms of the assumed interconnect topology, using an H-tree structure for routing within each hierarchy implies that the interconnections within individual tiles adhere to an H-tree topology. Each layer functions as an individual pipeline stage, and the system clock cycle for the pipeline is determined by the longest latency observed among all the layers. This setup implies that the layers are sequentially processed in a pipelined manner, where the output of one layer serves as the input to the subsequent layer. The framework employs an off-chip offloading model where a portion of the neural network layers is loaded into on-chip memory arrays while the remaining layers are stored in off-chip memory. Offloading entails transferring these layers between on-chip and off-chip memory, resulting in potential performance and power overheads. We have yet to develop a comprehensive analysis of the performance and power overheads associated with offloading, but we are actively working on it to obtain all the necessary details. Nevertheless, the offloading process would introduce additional latency and energy consumption due to the data transfer between on-chip and off-chip memory.

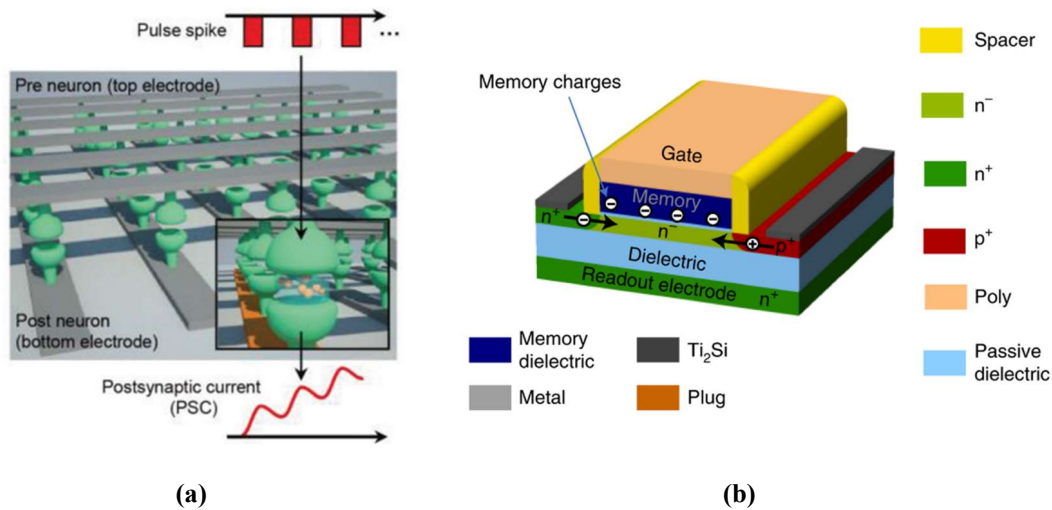

(a) (b)

**Fig. 3.3 Device structure (a) TiOx-based memristor device (b) Si-based memcapacitor device**

NeuroSim efficiently calculates the weight-matrix size for each layer in the pre-defined network structure using the weight mapping method. The process of iteratively reducing the matrix size involves the following sequence of actions: Initially, the tile size is configured to accommodate the largest weight matrix among all layers. Subsequently, the framework calculates memory utilization by dividing the memory mapped by synaptic weights by the total chip memory. The tile size is then gradually decreased while monitoring its impact on memory utilization. The aim is to achieve optimal memory utilization. This stepwise reduction in tile size contributes to refining memory allocation for improved efficiency. The weights are programmed using the conductance of the memory devices. When input vectors are encoded using read voltage signals, the weighted sum operation is performed in parallel, resulting in currents at the end of each column. The read voltage applied at the input of transmission gates passes through the WL, and the parallel readout of weighted sums occurs through the BL. In cases where input vectors are larger than 1 bit, encoding necessitates multiple clock cycles. The network employs a unit cell arrangement, and for encoding the inputs, 8 bits are used. This choice is influenced by the nature of the CIFAR-10 dataset used in the experiments, which comprises 32x32x3=3072 input features. Utilizing lower bit resolutions for encoding inputs would likely result in a degradation of network accuracy. In cases where negative inputs are encountered, they are encoded using the two's complement representation. In this representation, a negative value is represented by taking the complement of its positive counterpart and adding 1. Subsequently, the corresponding output is decoded using the same two's complement representation method.

Fig. 3.2 (a) and (b) depicts a typical design of a memristive and memcapacitive crossbar array utilized for realizing VMM. The design comprises a 2-D array of synaptic devices, digital-to-analog converters (DACs), analog-to-digital converters (ADCs), and write peripheral circuitry. As proposed in the framework [10], the NeuroSim core is enveloped by Python and PyTorch, allowing for the facilitation of flexible network topologies. The model used is 8 layer VGG network for CIFAR-10. However, the framework also supports larger models such as ResNet, AlexNet, GoogleNet, or users have the option to define arbitrary CNN topologies.

### 3.1.2 Memristor and Memcapacitor based VMM

Fig. 3.2 (a) and (b) depicts a typical design of a memristive and memcapacitive crossbar array utilized for realizing VMM. This framework comprises two parts: one implemented in Python and the other in PyTorch CUDA. For the evaluation of metrics such as nonlinearity, asymmetry, device-to-device variation, cycle-to-cycle variation, IMC area, latency, and energy. To assess the area, latency, dynamic energy, and leakage associated with interconnects, we assume that routing among modules within each hierarchy follows an H-tree structure. The latency and energy breakdown analysis reveals that, due to substantial on-chip data transfer, the primary bottlenecks are buffer latency and DRAM energy consumption. The estimated training dynamic energy per epoch amounts to 108.36 J, while the training latency per epoch is calculated at 104.31 sec. Although the paper mentions using mem-elements-based IMC arrays for weight gradient computation, it does not explicitly specify the exact on-chip storage capacity required for storing intermediate results or facilitating routing to multiple arrays. The size of the on-chip storage would be contingent upon factors such as the neural network's dimensions and the nature of the operations being conducted.

We consider six convolutional layers and two fully connected layers, each serving as dedicated computation units designed specifically for weighted sum and weight update operation. In the forward convolutional layer, the analog weights are initially mapped to memductance, with the line resistance serving as the memductance weight. The input data is then fed from the input layer and travels forward through a series of weighted sum operations and neuron activation functions until reaching the output layer. In the fully-connect layer, a similar process occurs, where the analog weights are first mapped to memductance. The VMM is performed on the input vector with the crossbar array assigned weights. During back-propagation, the error is propagated backward from the output layer to adjust the weights of each layer, minimizing the prediction error.

In the backpropagation step, there are two VMM steps. First, multiplying the weight matrix with gradients. It can be inferred that the second Vector-Matrix Multiplication (VMM) is executed by retrieving the activations from off-chip memory and then multiplying them with the gradients. When computing weight gradients, activations are fetched from off-chip memory and conveyed to on-chip buffers before reaching the weight gradient computation units. This implies that the activations are present in on-chip buffers and can thus be employed for the second VMM operation.

The TiOx memristive device characteristics encompass operational yield and uniformity, symmetrical analog switching, functional stability, and adjustable learning rates. The TiOx memristor array achieves a remarkable operational yield exceeding 99%, displaying exceptional uniformity in its switching threshold. Its symmetrical analog switching behavior enables both conductance potentiation and depression, essential for implementing synaptic functions in artificial neural networks. Notably, the device exhibits high functional stability, maintaining repeatability over 3000 programming cycles and remaining operational for six months. In essence, the TiOx memristive device showcases reliable symmetrical analog switching traits, operational uniformity, and functional stability, rendering it a promising candidate for effective in situ training within neuromorphic computing systems [1]. Fig. 3.3 (a) illustrates the device structure and image of a TiOx memristor [1]. This memristor features a crossbar array with nodes based on TiOx. The individual memristor cells are positioned at the intersections of Al electrode lines on a glass substrate, with each Al electrode line having a width of 100 μm.

The characteristics of the Si memcapacitive device, as described in [11], make it well-suited for neuromorphic computing applications. Notable features include a high dynamic range, which enables precise analog signal processing, and low power operation through adiabatic charging, enhancing energy efficiency. The device's scalability down to around 45 nm and its crossbar array architecture further support its integration into compact and energy-efficient neuromorphic systems. This architecture facilitates parallel multiply-accumulate (MAC) operations, ideal for neural network training and pattern recognition tasks. Overall, the Si memcapacitive device's dynamic range, low power operation, scalability, and crossbar array structure position it as a promising choice for energy-efficient neuromorphic computing systems [11]. The Si memcapacitive device comprises a layered structure with a gate electrode, shielding layer, and readout electrode. The gate electrode applies input signals, and the readout electrode reads accumulated charge. The shielding layer between them significantly affects capacitance modulation. The device structure includes a lateral pin junction and electron and hole injection, depicted in Fig. 3.3 (b).

The expected relationship between weight increase long-term potentiation, (LTP) and weight decrease long-term depression, (LTD) should be linearly dependent on the number of write pulses. However, real-world devices, as described in existing literature, often deviate from this ideal trajectory. In practice, the memductance tends to undergo rapid changes during the initial stages of LTP and LTD, eventually reaching a saturation point, as depicted in Fig. 3.2 (c) and (d) for TiOx-based memristive and Si-based memcapacitive devices, respectively.

For the TiOx-based memristive VMM [1] and, XLTP and XLTD represent the conductance values for LTP and LTD, respectively and the following are the equations:

$$X_{LTP} = B\left(1 - e^{\left(\frac{P}{A}\right)}\right) + X_{min} \tag{3.1}$$

$$X_{LTD} = -B\left(1 - e^{\left(\frac{P - P_{max}}{A}\right)}\right) + X_{max} \tag{3.2}$$

$$B = X_{max} - X_{min} \Big/ 1 - e^{\left(\frac{-P_{max}}{A}\right)} \tag{3.3}$$

Similarly, the equations mentioned above can be applied to the Si-based memcapacitive VMM [11]. In this scenario, XLTP and XLTD denote the capacitance values associated with LTP and LTD, respectively.

The parameters Xmax, Xmin, and Pmax are directly obtained from experimental hardware data and correspond to the maximum conductance and capacitance, minimum conductance and capacitance, and the maximum pulse number needed to switch the device between its minimum and maximum conductance states. The parameter A governs the nonlinear behavior of weight update and can be either positive (blue) or negative (red). In Fig. 3.2 (c) and (d), both LTP and LTD have the same magnitude but opposite signs for the parameter A. B, on the other hand, is a function of A designed to fit the functions within the range of Xmax, Xmin, and Pmax.

## 3.2 In-memory Computing with Non-Idealities

This section focuses on investigating nonidealities in memristive and memcapacitive crossbars and analyzing how they affect VMM.

### 3.2.1 Crossbar and Device Non-Idealities

Due to fabrication imperfections, non-ideal behaviors are observed in memristor and memcapacitor devices. These include variations in conductance, capacitance, device-to-device (D2D), and cycle-to-cycle (C2C), as well as nonlinearity and programming failure [12]. Consequently, it is crucial to consider nonuniformly distributed levels and conductance variations in the simulation of DNN. Devices are assigned to different levels based on the conductance and capacitance distribution in the crossbar array to assess the degradation of training accuracy under nonideal properties. The nonlinearity and asymmetry model can be represented by equations 1 to 3. The parameter A determines the degree of nonlinearity in weight update, with a value range of $(0, +\infty)$, where smaller A values indicate a more nonlinear weight update behavior. The device's conductance is programmed from a high resistance state (HRS) to a low resistance state (LRS) and shown in Fig. 3.4 for the memristive device. It is crucial to emphasize that while the failure mask undergoes updates in each programming cycle, the stuck mask remains unchanged throughout both training and inference. This is due to the inability to fix stuck devices, resulting in their fixed position after array testing.

In the weight update process, D2D variation leads to varying nonlinearities in different synaptic devices. To create a behavior model, we randomly generate the nonlinearity factors for different synaptic weights, using a standard deviation ($\sigma$) with respect to the mean nonlinearity value ($\mu$). The results depicted in Fig. 3.5 (a) highlight the significant impact of device variation, showcasing the remarkable accuracy maintenance achieved by our proposed method using both devices. To investigate the effects of C2C variation, we created a behavior model similar to the one used for device-to-device variation. C2C variation pertains to the variability in conductance change with each programming pulse. As shown in Fig. 3.5 (b) C2C variation does not degrade the performance of the system. Therefore, we can represent the cycle-to-cycle variation standard deviation ($\sigma$) as a percentage of the entire weight range.

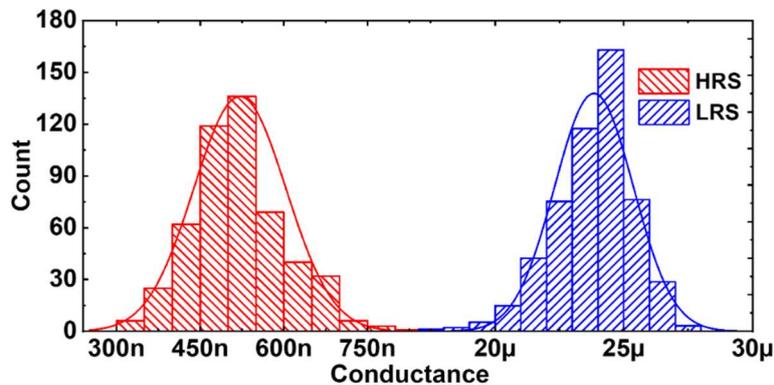

**Fig. 3.4 Distribution of the TiOx memristor conductance in the HRS and LRS**

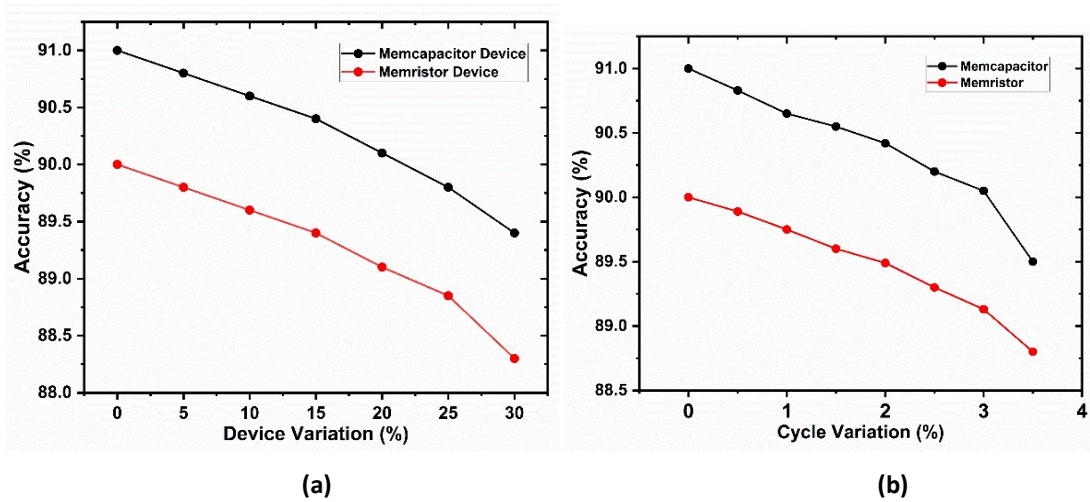

**Fig. 3.5 Training accuracy comparison of non-idealities (a) different device-to-device memductance variation (b) different cycle-to-cycle variation**

In an ideal scenario, currents in resistive crossbars should flow from left to right along the rows and from top to bottom through the columns. Nonidealities, including wire resistances, cause variations in the actual voltage across the memristor and memcapacitor VMM accelerator, resulting in a lower voltage than the theoretical value. This reduction is due to the accumulated voltage drop on the connecting traces and sneak pathways [13]. The presence of line resistance and sneak paths impacts the training accuracy of the model, as depicted from Fig. 3.6. In our results, we observed that the training accuracy of both VMM approaches is more significantly influenced by the line resistance than by the variations between individual devices and cycles. This suggests that the impact of line resistance plays a more prominent role in affecting the training accuracy compared to the inherent variability between devices and cycles.

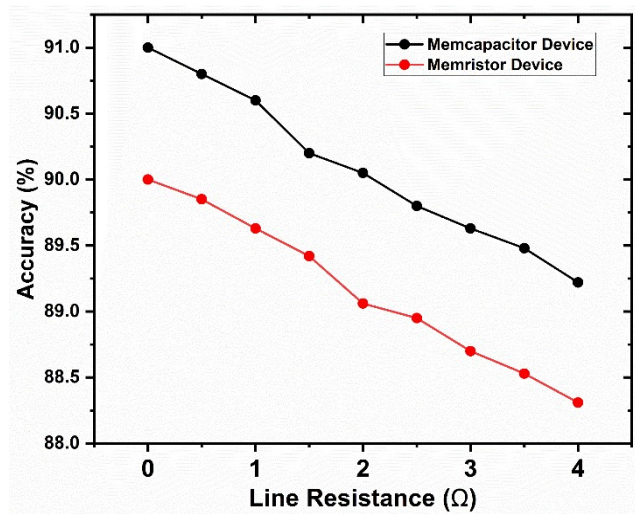

**Fig. 3.6 The accuracy is influenced by the line resistance and the sneak paths**

### 3.3.1 Peripheral Circuits

The VMM employs various peripheral circuit modules, including a switch matrix, multiplexer, adder, shift register, driver, and ADC [12]. In this framework, these peripheral circuits are designed using transistor parameters directly extracted from the TSMC 22-nm PDK, as shown in Fig. 3.7, and specifically set in the NeuroSim transistor library. These parameters encompass device W/L, supply voltage (VDD), threshold voltage ($V_{TH}$), gate and parasitic capacitance, and NMOS/PMOS on/off current density. By utilizing these parameters, the area and intrinsic RC/power model of standard logic gates can be analytically calculated using specific formulas, as discussed in prior works. This enables the estimation of performance metrics for each sub-circuit. The transistor W/L for the ADC, multiplexer, switch matrix, and drivers are predefined based on the required drivability, while the transistor W/L for other logic gates is set at a fixed size. The capacitances at the logic gate level are also improved, and their transistors' sizing is known. This allows for the calculation of τ = RC and CVDD2 to estimate module delay and dynamic energy consumption.

Switch matrices are vital components that facilitate fully parallel voltage input to the rows or columns of the array. These matrices are composed of transmission gates connected to all the bit lines (BLs), with their control signals stored in registers. The traditional crossbar word line (WL) decoder has been modified to activate all the WLs, making all the transistors transparent for weighted sum. This enhanced crossbar WL decoder integrates follower circuits into each output row of the conventional decoder. Additionally, a multiplexer is employed to distribute the read periphery circuits among the synaptic array columns, optimizing the utilization of resources as the array cell size is significantly smaller than the size of read periphery circuits. Placing all the read periphery circuits at the edge of the array would not be area efficient. Hence, the multiplexer efficiently addresses this challenge.

We have incorporated quantization noise for the ADC. To address the potential effects of ADC truncation on partial sums, we adopt a nonlinear quantization approach utilizing several quantization edges, each indicative of different levels of ADC precision. These edges are determined based on the distribution pattern of partial sums. Currently, we have not incorporated a read noise model. However, we intend to incorporate such a model in future endeavors. The CMOS transistor parameters are extracted from TSMC's PDK and integrated into the Framework. Components like current sense amplifiers (CSA), voltage sense amplifiers (VSA), level shifters, and switch matrices are realized using these CMOS parameters. However, achieving the same training accuracy in the hardware implementation may be challenging due to factors like fabrication mismatches, ambient noise, and other variables that can impact system performance.

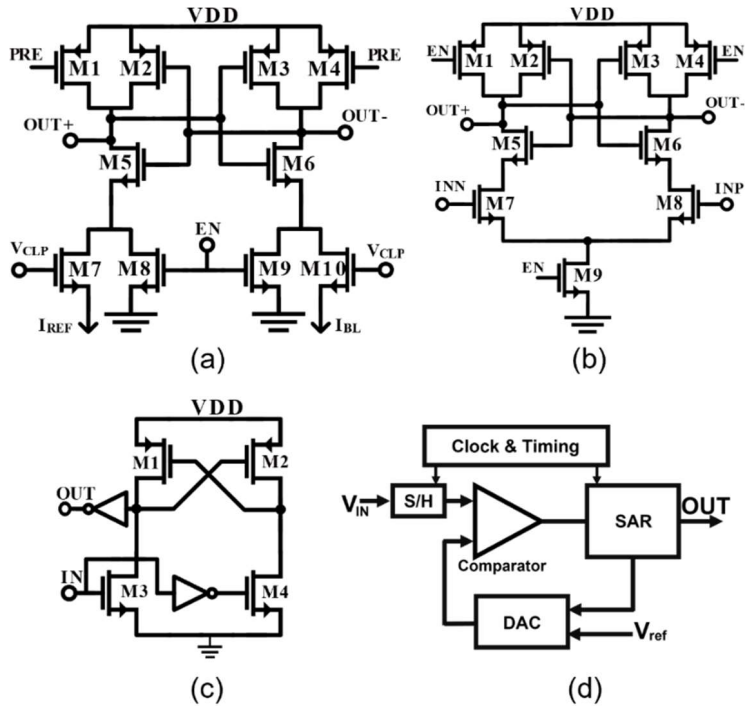

**Fig. 3.7** Transistor level used in the framework (a) current sense amplifier (CSA), (b) voltage sense amplifier (VSA), (c) level shifter, and (d) successive approximation register (SAR) ADC

The current sense amplifier (CSA) as shown in Fig. 3.7 (a) serves to amplify and convert small current signals into voltage signals, a critical function in precise analog-to-digital conversions within flash-ADCs. Fig. 3.7 (b) shows the voltage sense amplifier (VSA) plays a pivotal role by amplifying and converting minute voltage signals into digital outputs, essential for accurate conversion of analog voltages in ADCs, particularly in flash-ADCs. Additionally, a level shifter functions as a peripheral module facilitating translation of signal voltage levels across different logic domains as shown in Fig. 3.7 (c). This ensures seamless communication and signal compatibility among various parts of integrated circuits operating at diverse voltage levels. Lastly, in Fig. 3.7 (d) the successive approximation register (SAR) ADC operates by employing a binary search algorithm to determine the digital representation of analog input signals. Through iterative adjustment of the digital output, the SAR ADC converges to the closest digital representation of the input signal, making it suitable for various applications due to its moderate conversion speed and relatively low power consumption. The active blocks in the design operated at 1.1 V for VDD. Level shifters were incorporated into the design, particularly for the WL (Word Line), BL (Bit Line), and SL (Sense Line) signals within the crossbar array. These level shifters were employed to convert the voltage levels of these signals to the necessary levels essential for the correct functioning of the crossbar array.

To extract and process partial sums for subsequent logic modules, a group of flash-ADCs with multilevel successive approximation (S/A) using varying references is employed at the end of the synaptic lines (SLs) to produce digital outputs. In the simulator, a conventional current-sense-amplifier (CSA) based on transistor is utilized as the unit circuit module for building the multilevel S/A, as depicted in Fig. 3.7 (a). At the bottom of the synaptic core, an adder and shift register pair are utilized to execute shift and addition operations on the weighted sum result during each input vector bit cycle, resulting in the final weighted sum. The bit-width of the adder and shift register may need to be extended based on the precision requirements of the input vector.

# Chapter 4

## Proposed CMOS Meminductor for Neural Network

In this chapter, a type floating meminductor emulator setup is created in this study. The pinched hysteresis loop at various frequency ranges, the impact of temperature fluctuation on the circuit, and the change of meminductance according to the applied voltage signal pulses are all well discussed. Utilizing the cadence virtuoso environment tool and the characteristics of the 180 nm complementary metal oxide semiconductor (CMOS) technology, simulation results are produced. The suggested circuit is comparatively easier, electronically adjustable, lower number of transistors count and VLSI design compatible. In addition, the proposed work's applicability are verified through the design of a neuromorphic circuit for amoeba behavior, and vector-matrix multiplication (VMM) for convolutional neural network (CNN).

**4.1 Proposed Circuit Design**

### 4.1.1 Operational Transconductance Amplifier

In Fig. 4.1 (a), circuit symbols for an OTA [14] are shown, both of which have electronically controllable transconductance gain (gm) and high input and output impedance terminals for voltage and current, respectively. The MOSFETs implementation of OTA is illustrated in Fig. 4.1 (b). The output of OTA for an input Vp and Vn is expressed as

$$\left.\begin{array}{l} I_{O_i^+} = + g_{mi}(V_{p_i} - V_{n_i}) \\ I_{O_i^-} = - g_{mi}(V_{p_i} - V_{n_i}) \end{array}\right\} \quad (4.1)$$

where gm is the transconductance gain of the OTA. The routine analysis results in the expression of gm as

$$g_{mi} = \frac{K}{\sqrt{2}}(V_{bi} - V_{ss} - 2V_t) \quad (4.2)$$

Where "i" is the OTA number, "K" is a parameter of the MOS device given by

$$K = \mu_n C_{OX} \frac{W}{L} \quad (4.3)$$

The W denotes the channel width, L is the channel length, μn is the mobility of the carrier, and $C_{OX}$ is the oxide capacitance per unit area of the MOSFET.

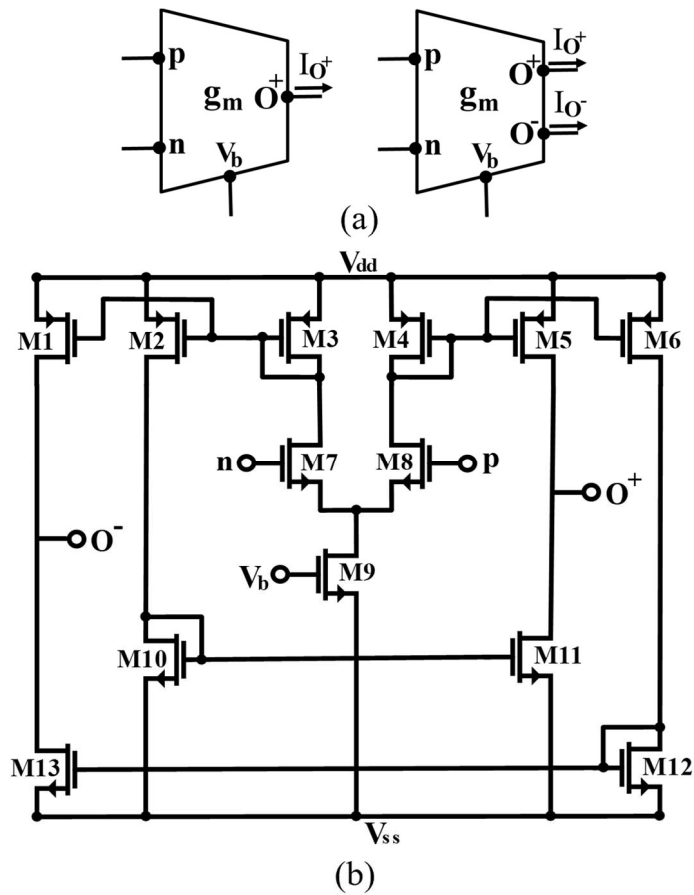

**Fig. 4.1 OTA (a) Symbol representation, and (b) MOSFETs realization**

### 4.1.2 Proposed Meminductor

The meminductor is also designed, as shown in Fig. 4.2. Through normal analysis, it generates the following set of equations:

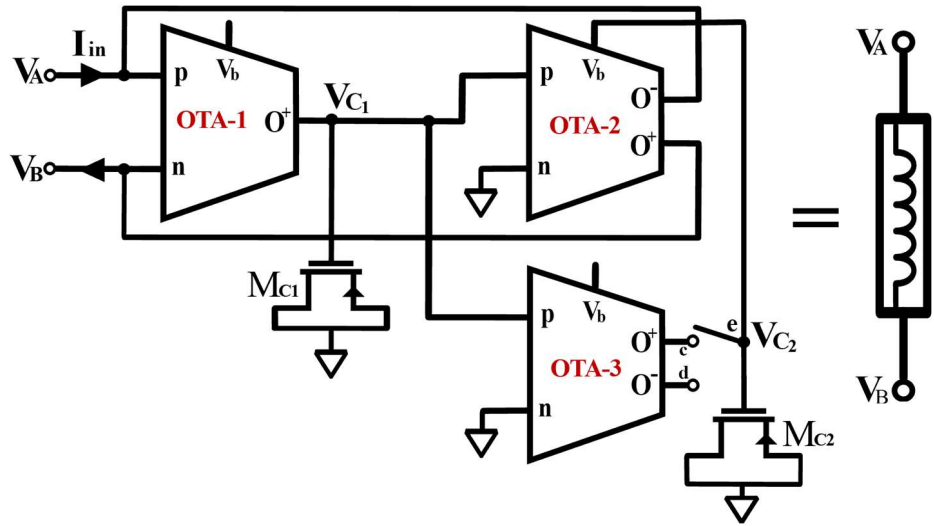

Fig. 4.2 Proposed meminductor emulator circuit

| S. no. | Pins Connections | Mode of operation |
|--------|------------------|-------------------|
| 1 | c-e | Decremental |
| 2 | d-e | Incremental |

Table 4.1 Connection Topology for pins for two modes

Considering the decremental meminductor emulator, the voltage at $M_{C1}$ is $V_{C1}$, at $M_{C2}$ voltage is $V_{C2}$ and Vin = $V_A$ - $V_B$, and the input current ($I_{in}$) terminal is "O-" of OTA-2.

Routine analysis of the design using Eq. 4.1 is given as:

$$I_{O_2} = I_{in}(t) = - g_{m2}(V_{p_1} - V_{n_1})$$
$$I_{in}(t) = - g_{m2}(V_{C_1}) \quad (4.4)$$

$$I_{O_3^+} = g_{m3} V_{C_1}(t) \tag{4.5}$$

$$I_{O_1^+} = g_{m1} V_{in}(t) \tag{4.6}$$

$$V_{C_1} = \frac{1}{M_{C1}} \int I_{O_1^+} \, dt \tag{4.7}$$

$$V_{b_2} = V_{C_2} = \frac{1}{M_{C2}} \int I_{O_3^+} \, dt \tag{4.8}$$

Substituting Eq. 4.6 into 4.8, we get:

$$V_{C_1} = \frac{1}{M_{C1}} \int g_{m1} V_{in}(t) \, dt = \frac{g_{m1} \phi(t)}{M_{C1}} \tag{4.9}$$

where, $\int V_{in}(t) \, dt = \phi(t)$ and putting eq. 4.5 into 4.7 to get $V_{C2}$.

$$V_{b_2} = V_{C_2} = \frac{1}{M_{C2}} \int g_{m2} V_{C_1} \, dt \tag{4.10}$$

$$V_{b_2} = V_{C_2} = \frac{g_{m1} g_{m3}}{M_{C1} M_{C2}} \rho(t) \tag{4.11}$$

Here, $\rho(t) = \int \phi(t) \, dt$ and bias voltage ($V_{b2}$) is equal to $V_{C2}$ of OTA-2 and using eq. (4.2) $g_{m2}$ is given by:

$$\begin{aligned} g_{m2} &= K(V_{C_2} - V_{ss} - V_t) \\ g_{m2} &= K\left(\frac{g_{m1} g_{m3} \rho(t)}{M_{C1} M_{C2}} - V_{ss} - V_t\right) \end{aligned} \tag{4.12}$$

Substituting equation (4.11) and (4.12) into (4.3) the input current can be expressed as

$$I_{in}(t) = -K\left(\frac{g_{m1}g_{m3}\rho(t)}{M_{C1}M_{C2}} - V_{ss} - V_t\right)\frac{g_{m1}\phi(t)}{M_{C1}}$$
$$\Rightarrow I_{in}(t) = K\left(V_{ss} + V_t - \frac{g_{m1}g_{m3}\rho(t)}{M_{C1}M_{C2}}\right)\frac{g_{m1}\phi(t)}{M_{C1}} \quad (4.13)$$

As a consequence, the corresponding inverse meminductance can be calculated as follows:

$$M^{-1} = \frac{I_{in}(t)}{\phi(t)} = \frac{Kg_{m1}}{M_{C1}}\left(V_{ss} + V_t - \frac{g_{m1}g_{m3}\rho(t)}{M_{C1}M_{C2}}\right) \quad (4.14)$$

The suggested design depicts a decremental meminductor model whose starting meminductance value is [((Kg$_{m1}$)/M$_{C1}$)(Vss+Vt)] as can be seen from the preceding equation, and the rate of transition of inductance is confirmed by [(Kg$_{m3}$(gm1)$^2$)/(( M$_{C1}$)$^2$ M$_{C2}$)]. whereas gm1, gm2, and gm3 transconductance of OTA respectively.

Similarly, the proposed incremental meminductor emulator may be achieved by modifying the circuit's switch as illustrated in Table 4.1:

$$M^{-1} = \frac{I_{in}(t)}{\phi(t)} = \frac{Kg_{m1}}{M_{C1}}\left(V_{ss} + V_t + \frac{g_{m1}g_{m3}\rho(t)}{M_{C1}M_{C2}}\right) \quad (4.15)$$

In order to assess the frequency domain behavior of the proposed meminductor emulator, the proposed meminductor emulator will be simulated by a Vin = Vmsin(t) sinusoidal voltage, whereas amplitude (Vm), and frequency (f).

The meminductance of decremental and incremental meminductor emulators is dependent on the amplitude and operating frequency of the sinusoidal signal, according to (4.14) and (4.15). The equivalent rate of transition in meminductance due to external simulation may thus be calculated as follows:

$$M^{-1} = \frac{I_{in}(t)}{\phi(t)} = \frac{Kg_{m1}}{M_{C1}}(V_{ss}+V_t) \pm \frac{Kg_{m1}g_{m3}V_m}{M_{C1}M_{C2}}\cos\left(\omega t - \frac{\pi}{2}\right) \quad (4.16)$$

$$M^{-1} \approx \pm \frac{Kg_{m1}g_{m3}V_m}{\omega M_{C1}M_{C2}} = \frac{1}{\lambda f} \quad (4.17)$$

where (+) denotes an incremental meminductor mode and (-) denotes a decremental meminductor mode for the operator. And λ = [(2π M$_{C1}$ M$_{C2}$)/(Kg$_{m1}$g$_{m3}$Vm] is the time constant of the proposed meminductor emulator circuit. The following time-constant conditions must be met to achieve proper meminductor performance:

(a) When the frequency approaches infinity, the meminductance element disappears, according to (4.17).

(b) The essential criterion for achieving the optimum pinched hysteresis loop area is (λ = 1/f), and

(c) If (λ < 1/f), when λ constant is less than input signal frequency, then the pinched hysteresis loop may be lost.

## 4.2 Proposed Meminductor Application in Neuromorphic

### 4.2.1 Neuromorphic Circuit for Amoeba Behavior

The viability of the suggested meminductor is exemplified as an adaptable neuromorphic framework. The progression of the mem system discovers utility in multiple domains, including an advanced application involving neuromorphic structures, owing to its ability to emulate the synaptic connections between neurons in the brain [15]. Furthermore, a neural network based on memristors is employed to execute a Pavlovian experiment, showcasing an adaptive learning process [16]. In contemporary times, the unicellular organism known as amoeba has been subject to scrutiny, with efforts focused on emulating its unique traits. These endeavors aim to unravel the rudimentary processes of learning and comprehend the intricate behaviors exhibited by the brain [16]. Amoeba, being the most elementary form of unicellular life, has persisted since the inception of life on Earth. Amoeba showcases a remarkable sensitivity to temperature variations, displaying a noteworthy response by decelerating their movement whenever a decrease in temperature occurs. Moreover, it possesses the capacity to anticipate impending temperature drops based on previous instances, highlighting its ability to exhibit future predictions.

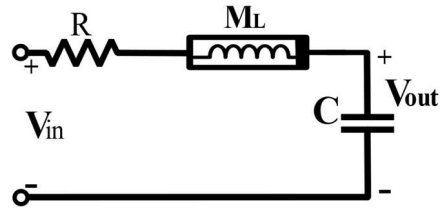

**Fig. 4.3 An RLC neuromorphic circuit using meminductor for amoeba behavior**

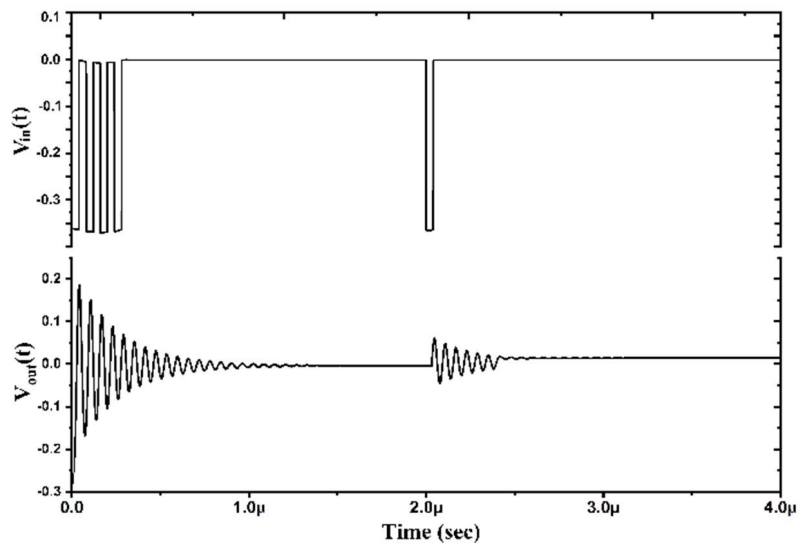

**Fig. 4.4 Simulation Result of neuromorphic circuit using meminductor for amoeba behavior**

This intriguing behavior necessitates the fulfilment of three essential requisites: 1) retention and recall of past experiences; 2) forecasting future events; and 3) precise timing of periodic occurrences. In, a circuit based on meminductors is introduced to replicate the adaptive behavior observed in amoebas. This circuit entails a straightforward RLC configuration, as depicted in Fig. 4.3. The same circuit is employed to validate the functionality of the proposed floating meminductor. Within the RLC circuit, the interplay between the capacitor and inductor generates oscillations at the resonant frequency, while the resistor (R) serves as an energy dissipation component that gradually dampens the oscillations over time. The input voltage, denoted as Vin(t), serves as a representation of the environmental temperature fluctuations, while the output voltage, Vout(t), corresponds to the locomotive speed of the amoeba, analogous to the variations in input temperature. The memsystem nature of the inductor allows for its value to be altered based on the historical voltage across the inductor. As a result, the inductor value adapts in accordance with the frequency of temperature variations (Vin(t)), eventually reaching a state of resonance when the frequency of the RLC circuit aligns with the input frequency. Fig. 4.4, exhibits the simulation results obtained using Cadence Virtuoso, showcasing both the input and output responses of a meminductor-based neuromorphic circuit. The circuit consists of specific component values, with R set at 1 kΩ and C at 10 pF. Observing Fig. 4.4, it becomes evident that with each instance of temperature drop, the corresponding output, representing the amoeba's locomotive speed, also experiences a decrease. However, it is notable that the output response exhibits a significant delay before accurately reflecting the locomotive response, which can be attributed to the phenomenon known as the "delayed switching effect" [17]. This effect refers to the temporal duration required for transitioning between different states within the mem-system, resulting in a noticeable time delay. Consequently, considering the application perspective, the proposed meminductor design proves to be well-suited for the intended purpose.

### 4.2.1 VMM Accelerator for CNN

The meminductor-based VMM accelerator is a novel hardware architecture designed for efficient and enabling parallel products. It leverages meminductors, which are memory devices capable of dynamically adjusting their inductance values based on voltage and current history. Employing a crossbar architecture, the accelerator enables parallel processing of multiple elements in the input vector and matrix. As data flows through the meminductor array, the inductance values of the meminductors adapt, optimizing the multiplication process. This adaptive behavior ensures efficient computation and adaptability to varying computational requirements [18]. The VMM accelerator, designed to enhance Convolutional Neural Networks (CNN), uses proposed meminductor technology. Illustrated in Fig. 4.6 (a), the VMM accelerator showcases a schematic representation that prominently features a 3 × 3 meminductor array. This meminductor device is meticulously implemented through the utilization of commercial ICs, specifically the CA3080, in tandem with capacitors, a detailed explanation of which can be found in Section III (E). The integration onto the PCB is facilitated via wire bonding. Each meminductor device boasts a trio of input channels for data writing, complemented by an additional three input channels dedicated to the reading of data from their respective modules.

The VMM accelerator's operational process unfolds as follows: The zedboard PMOD device serves as the source of digital signals, transmitting them to the input side of the crossbar arrays. These signals represent binary states of either '0' (0 V) or '1' (3.3 V). When these signals reach the meminductor arrays, they induce corresponding currents, which subsequently traverse a 1.5 kΩ resistor as 'R.' This flow of current results in the generation of a voltage range spanning from 0 V to 3.4 V. Fig. 4.6 (a), the resultant output voltage (as 'yi') from the resistor is then directed to the zedboard XADC, and it is stored after processing. For the implementation of the initial layer within the convolutional architecture, a total of five modules have been deployed. The experimental setup with the zedboard and VMM accelerator is further visualized in Fig. 4.7. Here are the equations for mapping weight coefficients (wij) onto meminductor-based VMM accelerator circuit in Fig. 4.6 (a).

$$S_{ij} = \frac{w_{ij} - w_{min}}{w_{max} - w_{min}}(S_{on} - S_{off}) + S_{off} \quad (4.18)$$

$$Y_j = R.I_j \quad (4.19)$$

where, R is the resistor, Sij is the meminductance of the device in the ith row and jth column, Yj are the output, and Ij are the current of the jth column.

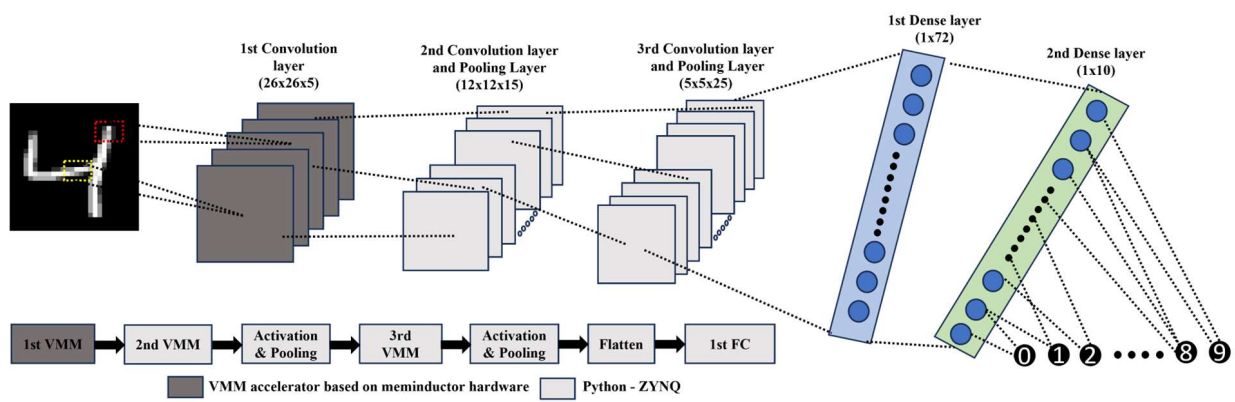

**Fig. 4.5 Structure of CNN implemented in software for classification of MNIST dataset**

To assess the performance of our meminductor-based VMM accelerator, we conducted an evaluation using a software-hardware co-designed CNN [18]. The CNN architecture comprises three convolutional layers and one fully connected layer, illustrated in Fig. 4.5. In the first, second, and third convolutional layers, we employed filter sizes of 5, 15, and 25, each using 3 × 3 kernel matrices. The hardware implementation was focused specifically on the first convolutional layer as shown in Fig. 4.8. Our network was designed with scalability in mind, organized into a total of 5 modules. Each module is equipped with 3 × 3 meminductor arrays. For the remaining convolutional layers, we utilized meminductor hardware specifications implemented in zedboard using python. Equation (4.18), describing meminductor behavior, was implemented in zedboard with parameter values such as K = 4.02e-4 and gm1 = gm3 = 967e-6. We kept the other parameters consistent with the provided simulation results.

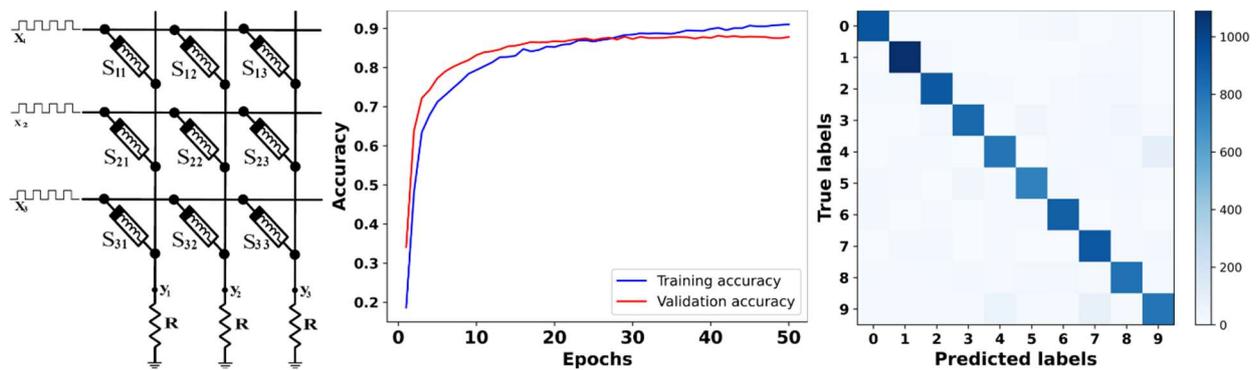

**Figure 4.6 Proposed meminductor-based (a) VMM accelerator. MNIST dataset (b) Training and Validation accuracy (c) Confusion metrices of classification results**

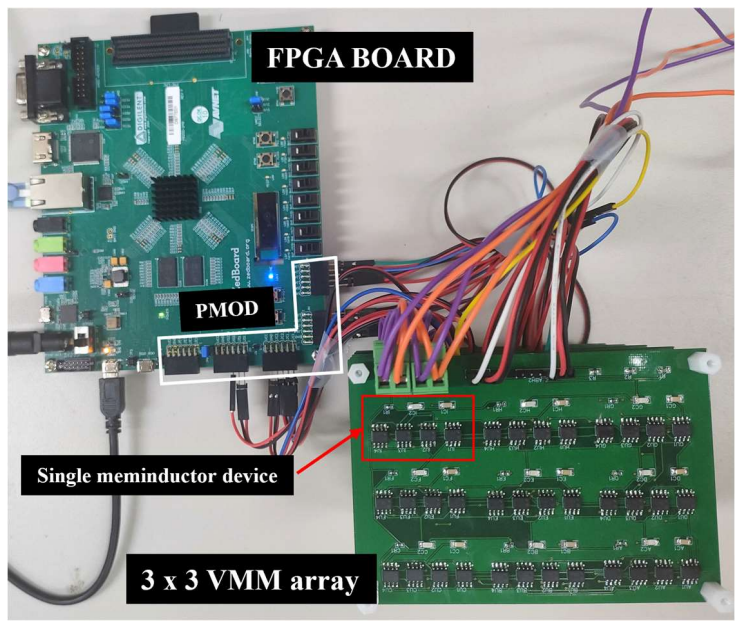

**Fig. 4.7 Photograph of experimental setup**

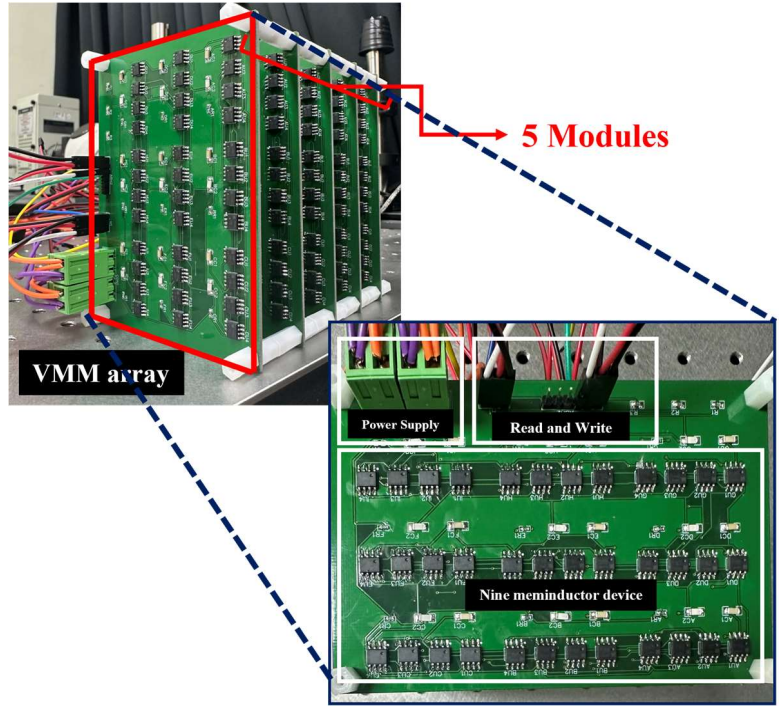

**Fig. 4.8 Schematic of the hardware-implemented convolution layer**

In the training phase, we utilized a dataset comprising 20,000 images from the MNIST dataset, each with dimensions of 28×28 pixels. The CNN model's training process is illustrated in Fig. 4.9. During each iteration of training, the network executed two crucial computations: forward propagation and backward propagation. Forward propagation in the CNN model involved a sequence of operations, including VMM, activation functions, and pooling. In the forward computations, input data traversed through the model's layers, incorporating the meminductor-based VMM accelerator. This accelerator optimized the multiplication process by dynamically adjusting meminductor inductance values based on voltage and current history. The resulting outputs underwent processing through activation functions, introducing non-linearity to enhance the model's ability to capture intricate patterns. Subsequently, pooling operations were performed to reduce the spatial dimensions of feature maps while retaining critical information. Backpropagation was implicitly handled by the Keras framework during model training using the model.fit() function. It calculated gradients, propagated them backward through the layers, and updated weights to optimize model performance. During testing, the trained network was evaluated using 10,000 test images from the MNIST dataset to assess its accuracy in classifying handwritten digits. To evaluate the proposed approach's performance, we analyzed training and test accuracy curves, reaching approximately 91.04% and 88.82%, respectively, as depicted in Fig. 4.6 (b). The classification results of the MNIST dataset were evaluated using confusion matrices, as shown in Fig. 4.6 (c). From an application perspective, our meminductor design aligns well with its intended purpose.

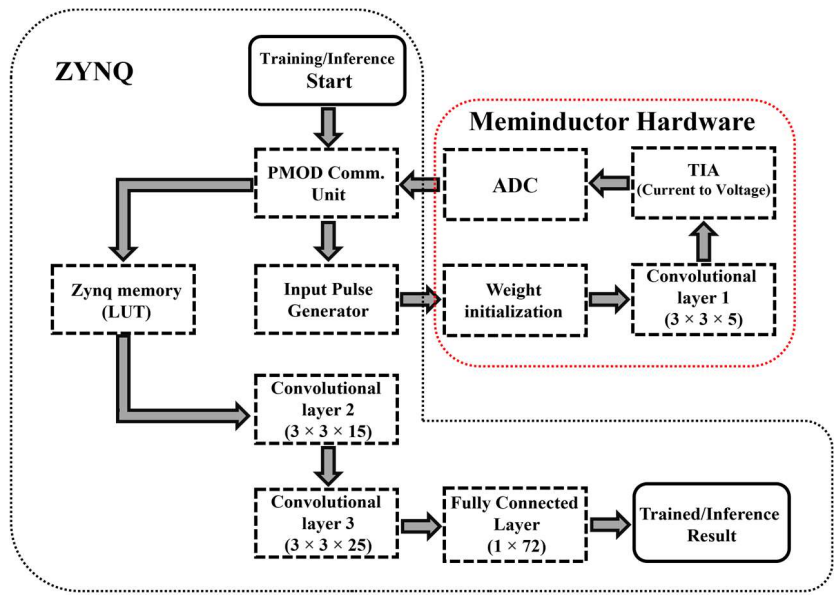

**Fig. 4.9 Flowchart of CNN model training**

# Chapter 5

# Simulation and Hardware Results

## 5.1 Hardware results of IMC based on Memristor and Memcapacitor

### 5.1.1 Framework Results

In this chapter, we examine detailed analysis for accuracy, area, throughput, and energy efficiency with our Python framework designed to evaluate large-scale memristive and memcapacitive VMM accelerator, and the hardware configuration for NeuroSim is in Table 5.1. We particularly emphasize the significance of on-state resistance (Ron), on-state capacitance (Con), and ADC precision in inference-only VMM accelerators, as these hardware factors play a crucial role in determining the accuracy and performance of the system. Our focus lies in analyzing data extracted from TiOx material memristive and Si material mem-capacitive devices based on the measurement results of 128 x 128 crossbar array, with specific attention to nonlinearity, crossbar sneak paths, asymmetry, device-to-device, and cycle-to-cycle variation for in-situ training. For benchmarking purposes, we utilize the 8-layer VGG network, which is shown in Fig. 5.1, with the CIFAR-10 dataset across various device technologies. In Table 5.1 each cell precision refers to the number of bits employed to represent the conductance of each individual memristor cell. For instance, a system utilizing 5-bit precision memristor or memcapacitor cells signifies that each memristor can express 32 distinct conductance levels. When we accumulate data from 128 of these cells, we effectively combine their conductance values to calculate the weighted sum of inputs. The utilization of 5-bit ADC precision in this scenario implies that the analog sum of these 128 cells is being discretized into 5-bit digital values. Nevertheless, it's crucial to recognize that the selection of ADC precision can vary based on the specific application's requirements and the desired level of accuracy. Increasing the bit precision can potentially enhance accuracy as well.

The framework also supports larger models as per the requirement. Simulation using the ResNet-34 network yielded training accuracies of approximately 86.14% for the memristor device and 85.81% for the memcapacitor device. These values were notably lower compared to the results from the VGG-8 network. However, for future iterations of the framework, we intend to include a wider range of networks and more complex datasets to enhance the comprehensiveness of our analysis. To determine whether the network is overfitting, we used a validation dataset to prevent overfitting, which is separate from the training dataset. We periodically monitored the model's performance on this validation dataset using metrics such as validation loss. The weights that minimized the loss or optimized the chosen metric were selected as the final trained weights. This approach ensured that the model did not overly adapt to the training dataset and could generalize well to new, unseen data.

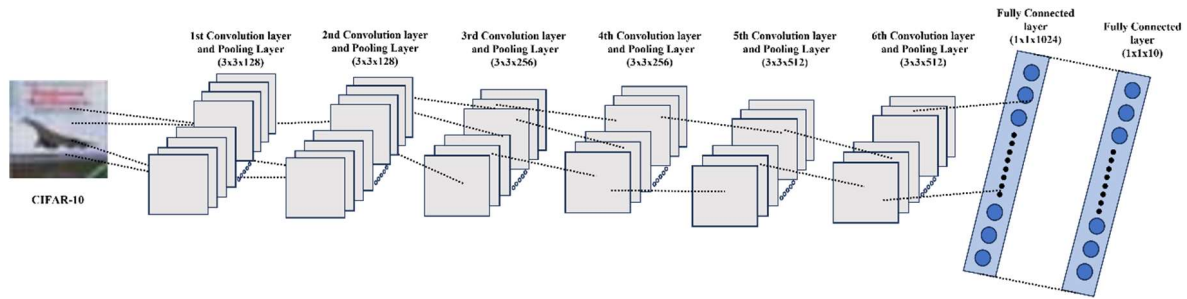

**Fig. 5.1 Proposed meminductor-based (a) VMM accelerator. MNIST dataset (b) Training and Validation accuracy (c) Confusion metrices of classification results**

| Configuration | Value |
|---|---|
| Operation mode | Conventional Parallel |
| ADC Precision | 5-bit |
| Crossbar Size | 128 x 128 |
| Clock Frequency | $10^9$ |
| Temperature | 300 K |
| Roff (Ω) | $25 \times 10^6$ |
| Ron (Ω) | $10^4$ |
| Coff (C) | $30 \times 10^{-12}$ |
| Con (C) | $2 \times 10^{-12}$ |
| Device Type | Memristor, Memcapacitor |
| Technology | 22 nm |

**Table 5.1 DNN NeuroSim experimental hardware configuration**

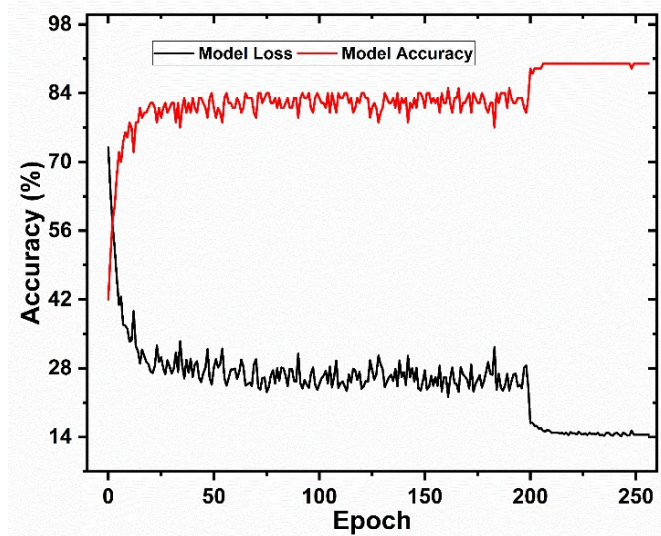

(a)

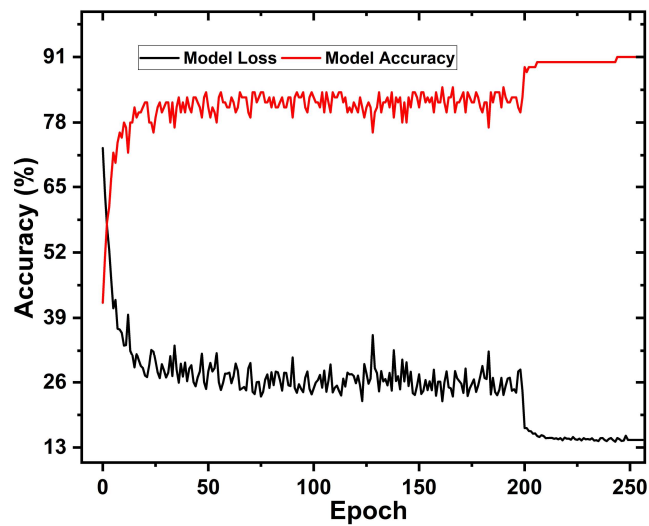

(b)

**Fig. 5.2 The training accuracy of the VMM accelerators with (a) memristor device, (b) memcapacitor device**

The proposed framework's performance is thoroughly assessed using system throughput (TOPS) and energy efficiency (TOPS/W). TOPS measures computational performance in terms of the number of trillion operations a system can perform per second. On the other hand, TOPS/W is the ratio of throughput to power consumption, indicating how efficiently the system performs computations per unit of power consumed. Higher TOPS and energy efficiency values are desirable for achieving more powerful and energy-efficient computing systems. Detailed information of the VMM accelerators on TOPS and TOPS/W can be found in Fig. 5.3 and Table 5.2. The proposed approach is also evaluated by analyzing the training accuracy curves of both VMM accelerators, achieving approximately 90.02% and 91.03% accuracy, as shown in Fig. 5.2.

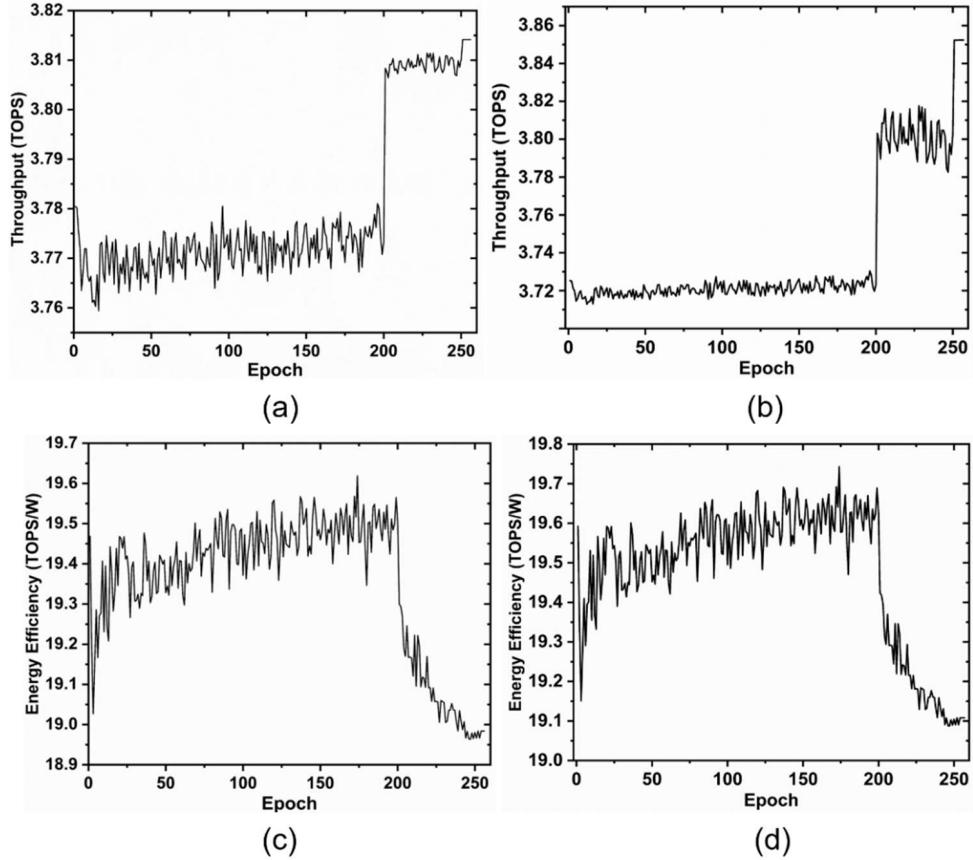

**Fig. 5.3 The framework was trained using the CIFAR-10 dataset for 256 epochs. Training throughput (TOPS) of VMM with (a) memristor device (b) memcapacitor device. Training Energy Efficiency (TOPS/W) of VMM with (c) memristor device (d) memcapacitor device**

### 5.1.1 Comparison with Prior Results

Table 5.2 provides a detailed summary of the proposed method's performance, including information about both VMM accelerators, and compares it with other recently published DNNs implemented with mem-elements crossbar systems, specifically for the CIFAR-10 image classification task. Notably, with a significantly smaller network size, the proposed framework achieves comparable inference accuracy with 8-layer VGG network using both memristive and memcapacitive devices. Table 5.2 provides a concise comparison between the proposed work and previously published works. The size of the crossbar in [19] and [21] is large, leading to impractical area consumption, whereas our proposed VMM accelerator achieves a smaller footprint compared to [22], [23], [24], and [25]. Additionally, our proposed work exhibits higher training accuracy compared to [20], [22], [24], and [25]. Furthermore, our approach demonstrates better performance in terms of throughput (TOPS) and energy efficiency (TOPS/W) compared to earlier works. Unlike [21], which utilizes digital ReRAM with certain limitations, our proposed work utilizes hardware-extracted data based on memristor and memcapacitor, thereby addressing nonidealities more effectively. The comparison demonstrates the superior effectiveness of the proposed VMM accelerator, showcasing its successful utilization of in-memory computing for on-chip training.

|  | This Work | | [19] | [20] | [21] | [22] | [23] | [24] |
|---|---|---|---|---|---|---|---|---|
| Device | Memristor (TiO$_X$) | Memcapacitor (Si) | Ag:a-Si | Digital ReRAM | EpiRAM | HZO FeFET | PCMO | AlO$_x$/HfO$_2$ |
| Network Structure | VGG-8 | | VGG | BNN-9 | VGG-8 | VGG-8 | VGG-8 | VGG-8 |
| Crossbar Size | 128 × 128 | | 128 × 128 | 128 × 128 | 784 × 300 | 128 × 128 | 128 × 128 | 128 × 128 |
| # of Conductance States | 32 | | 97 | - | 64 | 32 | 50 | 40 |
| ADC precision | 5-bit | | 6-bit | - | 6-bit | 6-bit | 6-bit | 6-bit |
| Weight/ Cell precision | 5-bit/ 1-bit | | 6-bit | - | 6-bit | 5-bit | 5-bit | 5-bit |
| Ron (Ω)/Con (C) | $10^4$/$2 \times 10^{12}$ | | $50 \times 10^6$ | $50 \times 10^3$ | $81 \times 10^3$ | $240 \times 10^3$ | $23 \times 10^6$ | $16.9 \times 10^6$ |
| On/Off Ratio | 10 | 10 | 12.5 | 16 | 50.2 | 100 | 6.84 | 4.43 |
| Device Variation (3σ/μ) | 30% | 30% | - | 12 % | - | - | - | - |
| Line resistance (Ω) | 0.5 | 0.5 | - | 0.1 | - | - | - | - |
| Area (mm$^2$) | 29.4 | 47.1 | 48.29 | 0.78 | 48.59 | 48.29 | 48.29 | 49.88 |
| Memory Utilization (%) | 88.59 % | 88.59 % | 88.59 % | - | 88.59 % | 88.59 % | 88.59 % | 88.59 % |
| Training Accuracy (%) | 90.02 % | 91.03 % | 49.00 % | 92 % | 85.00 % | 91.00 % | 56.00 % | 37.00 % |
| Training Throughput (TOPS) | 1.51 | 1.54 | 0.14 | 0.792 | 0.95 | 1.04 | 0.03 | 0.30 |
| Training Energy Efficiency (TOPS/W) | 2.10 | 2.32 | 2.00 | 176 | 2.00 | 2.01 | 2.00 | 1.98 |
| Training Peak Throughput (TOPS) | 3.81 | 3.85 | 0.16 | - | 2.68 | 3.57 | 0.03 | 0.38 |
| Training Peak Energy Efficiency (TOPS/W) | 18.98 | 19.11 | 20.54 | - | 20.11 | 20.57 | 20.50 | 17.27 |

Table 5.2 DNN NeuroSim experimental hardware configuration

## 5.2 Meminductor Results

The cadence virtuoso software's analog design environment tool checks the viability of the suggested meminductor emulator, and the proposed meminductor mathematical model is verified both in simulation mode and experimentally. The OTA block is designed using the CMOS structure shown in Fig. 4.1 (b) and is made using Taiwan Semiconductor Manufacturing Company (TSMC) 180 nm PDK technology with an aspect ratio (width/length) for pMOS and nMOS of 865 nm/360 nm. The supply voltages are VDD = −VSS = 0.9 V and bias voltage as Vb=− 0.12 V for OTA 1 and 3.

This study applies input sinusoidal signals having a peak voltage of 500 mV and frequency of 3 MHz to meminductor with 90◦ phase shift as depicted in Fig. 5.4 (a). MOS-capacitors are shown in Table 5.3, which shows the frequency range at different capacitor values. From the Fig. 5.4, when the input flux (ϕ(t)) is zero, the input current (Iin(t)) is zero as well [25] satisfying the constitutive relationship of the proposed emulator defined in Eq. 10. The ϕ(t) and the state variable (ρ(t)) determine its device characteristic. The pinched hysteresis curves of the proposed meminductor are shown in Fig. 5.4 (b). The suggested meminductor hysteresis curve also verifies the property of the ideal meminductor that passes through the origin. Note that the ϕ(t) is calculated as the voltage across $M_{C1}$, and the input current is the current flowing through the O- of OTA-2 terminal.

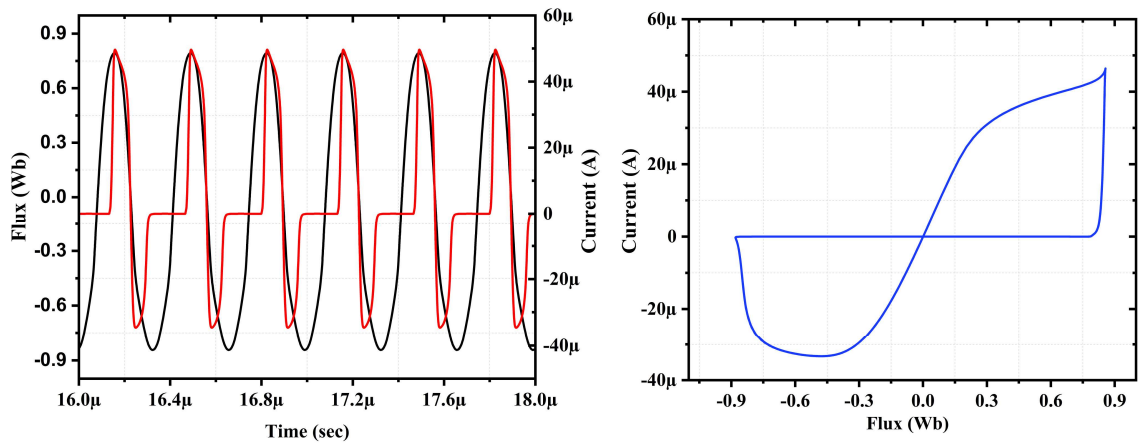

**Fig. 5.4 At 3 MHz, the response of a proposed meminductor. (a) meminductor transient response. (b) Pinched hysteresis curve for emulator**

| Frequency Range | $M_{C1}$ (Width) | $M_{C2}$ (Width) |
|---|---|---|
| 1 kHz to 9.99 kHz | 400 μm | 380 μm |
| 10 kHz to 1.99 MHz | 280 μm | 220 μm |
| 2 MHz to 6.99 MHz | 192 μm | 180 μm |
| 7 MHz to 11.9 MHz | 120 μm | 74 μm |
| 12 MHz to 17.9 MHz | 90 μm | 53 μm |
| 18 MHz to 25.9 MHz | 60 μm | 35 μm |
| 26 MHz to 39.9 MHz | 45 μm | 21 μm |
| 40 MHz to 60 MHz | 30 μm | 10 μm |
| Length (L) of both MOS- capacitors fixed to 2 μm | | |

**Table 5.3 MOS-Capacitors parameter and frequency range of proposed meminductor**

### 5.2.1 Parametric Analysis

The hysteresis curve's lobe area shrinks as frequency increases and turns linear at extremely high frequencies [25]. Fig. 5.5 illustrates the functioning of the proposed meminductor emulator tested at various frequencies. An important fingerprint of the meminductor is verified by this investigation [26]. To achieve a suitable hysteresis curve, the suggested meminductor's maximum working frequency is 60 MHz with different MOS-capacitors values, which is shown in Fig. 5.6.

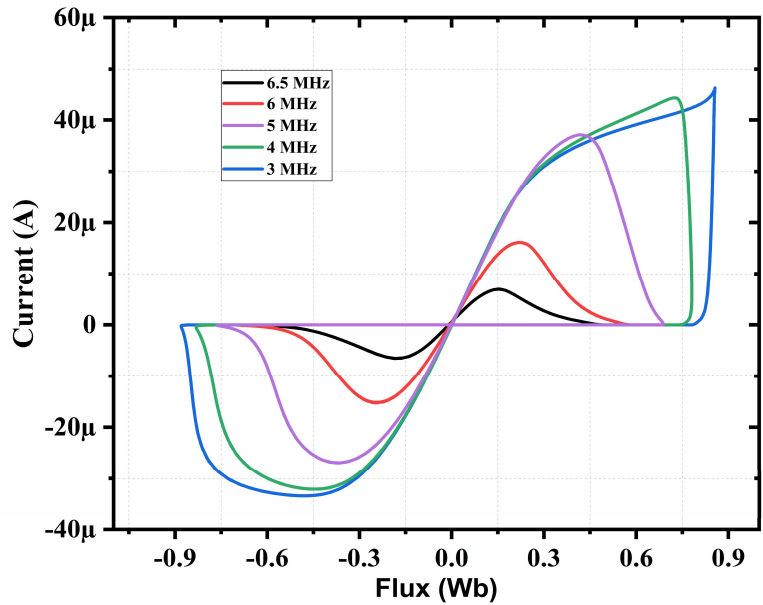

**Fig. 5.5 Meminductor's pinched hysteresis at various frequencies**

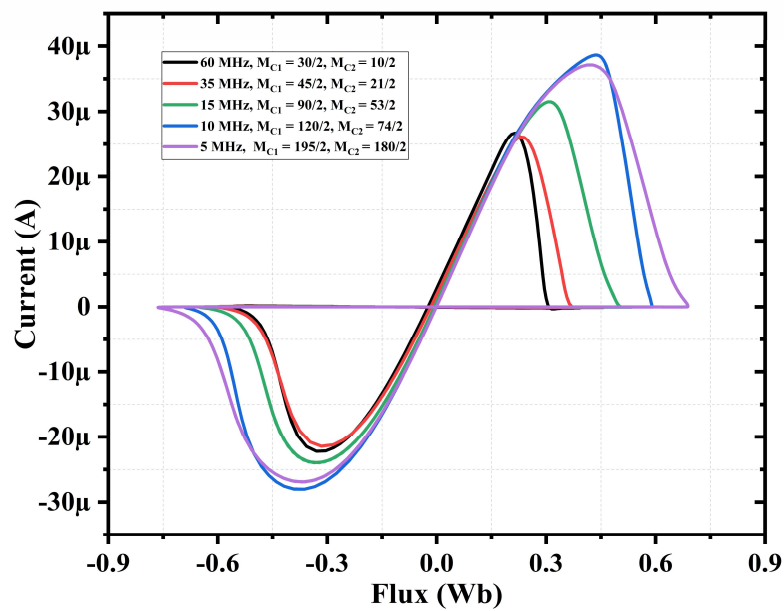

**Fig. 5.6 The proposed meminductor's pinched hysteresis at various frequencies and different MOS-Capacitors**

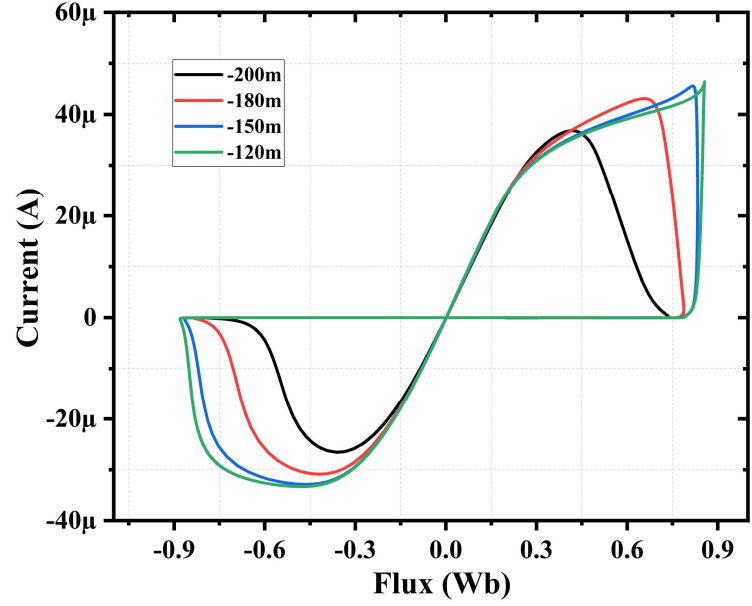

Fig. 5.7 Meminductor hysteresis loop at various bias voltages

It is observed that bias voltage Vb of OTA-1 and OTA-3 control transconductance gm1 and gm3, which can be used to tune meminductor characteristics. As shown in Fig. 5.7, the lobe area of the hysteresis curve is reduced as a result of the drop in gm1 and gm3 when the bias voltage is reduced.

The circuit design that has been proposed can be configured as either grounded or floating, allowing for it to be used in both serial and parallel connections. When multiple meminductors are connected in parallel, the overall meminductance is lower than that of a single meminductor. Conversely, when connected in series, the overall meminductance is greater than that of a single meminductor. This leads to a higher current in the parallel configuration and a lower current in the series configuration, as illustrated in Fig. 5.8. Transistors in the proposed meminductor cause deviations in meminductor characteristics as the modest change in bias voltage, process variations, and temperature variations. A detailed analysis of these variations is given as follows.

### 5.2.2 Temperature Variation

Temperature is an environmental factor affecting transistors, alongside saturation velocity, carrier mobility, and threshold voltage (Vth) [27]. These factors influence the drain current of transistors. As the recommended meminductor is based on a CMOS-based OTA circuit, it is important to investigate how temperature impacts the responses of the proposed meminductor. The pinched hysteresis curve of the proposed meminductor is analyzed at temperatures ranging from −40◦C to 40◦C, as illustrated in Fig. 5.9. It is evident that the lobe area of the hysteresis curve is larger for lower temperatures, and the area decreases as the temperature rises. The reason behind this phenomenon is that, at higher temperatures, the amount of current flowing through the circuit decreases, whereas at lower temperatures, the amount of current increases, thereby altering the hysteresis curve lobe area. While the temperature may also affect the value of resistors, the recommended meminductor design does not incorporate resistors, unlike the earlier design.

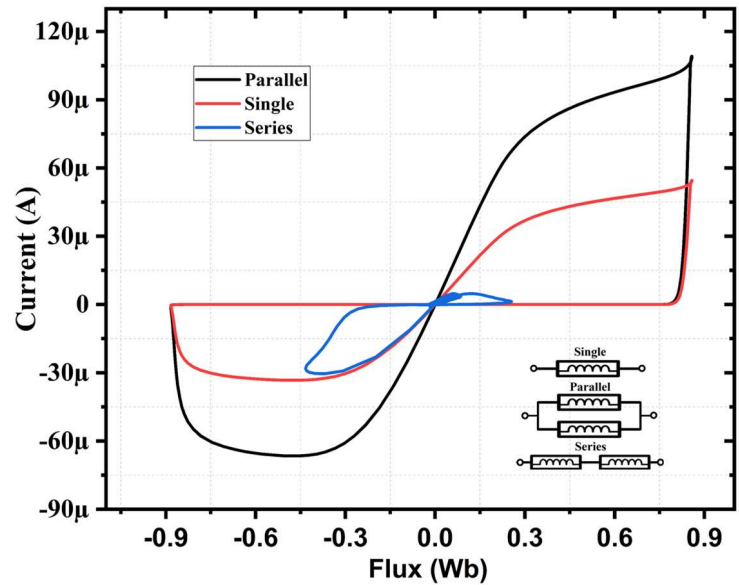

**Fig. 5.8 The input current-flux characteristics for single, parallel and series-connected meminductors**

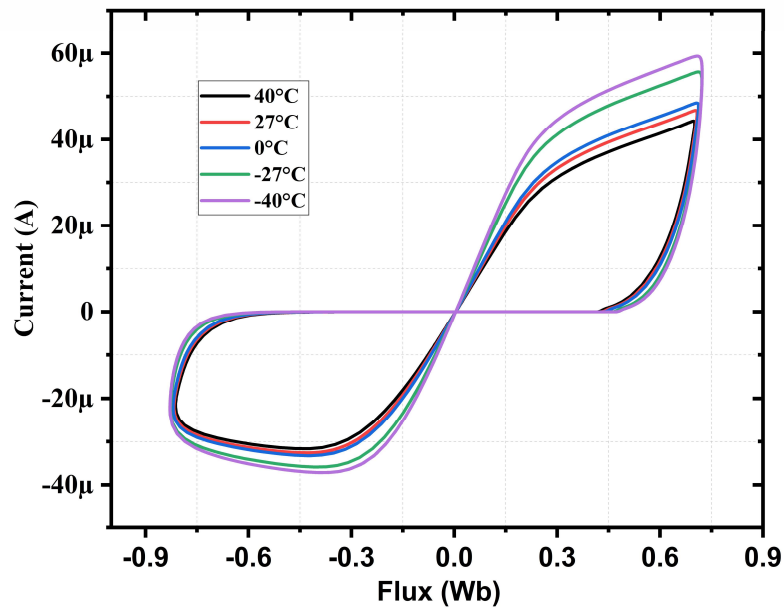

**Fig. 5.9 Flux vs Current curve of a proposed meminductor emulator at various temperatures**

### 5.2.3 Process Variation

When a design progresses toward monolithic integration, process variation is critical. Typical nMOS(N) and Typical pMOS(P) transistors (TT), fast N and fast P transistors (FF), slow N and slow P transistors (SS), slow N and fast P transistors (SF), and fast N and slow P transistors (FS) are all explored using the suggested meminductor. The investigation of process variation at 3 MHz as shown in Fig. 5.10. The lobe area of the hysteresis curve is the maximum at FF and minimum at SS, as expected. Despite the area variation in the hysteresis loop, the proposed meminductor circuits exhibit pinched hysteresis loops in all process corners. As shown in Fig. 5.11, the suggested design is described in terms of temperature and process corner fluctuations. It is clear that despite fluctuations in temperature and process corners, the behavior of the meminductor is maintained by producing hysteresis curve without any distortion.

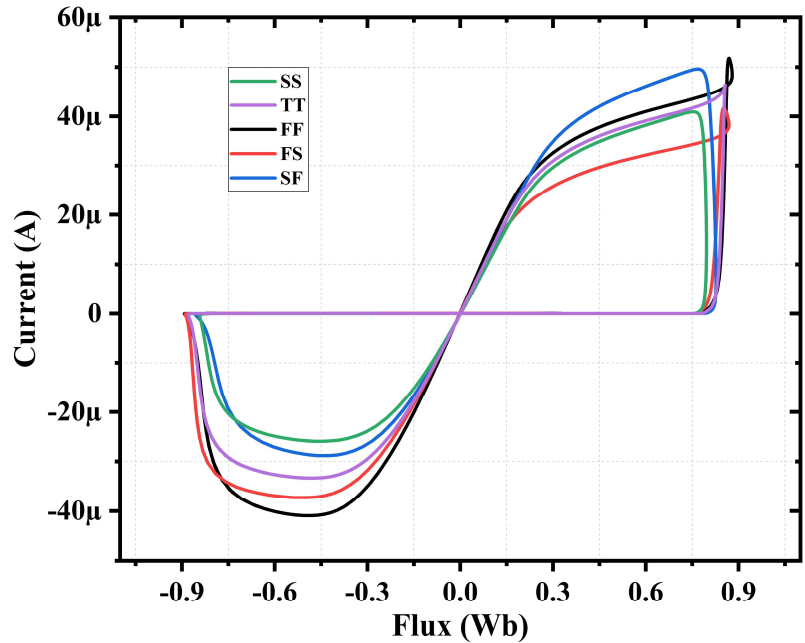

**Fig. 5.10 Variation in the hysteresis curve of meminductor at various process corners**

### 5.2.4 Post-layout Simulation

The proposed layout of the meminductor consists of a single-output OTA and two multi-output OTA integrated with two MOS capacitors. The layout of the meminductor is illustrated in Fig. 5.12, and its area utilization is 1690.21 µm2 (66.57 µm x 25.39 µm). To confirm the accuracy of the proposed meminductor emulator, pre- and post-layout analyses were performed using the Cadence Virtuoso framework and TSMC 180 nm PDK. The hysteresis curve from the pre and post-layout studies is shown in Fig. 5.13. It should be noted that the presence of parasitic components results in some variance between the post-layout hysteresis curve and the pre-layout hysteresis curve. However, this deviation was only a maximum of approximately 5%, which was still within acceptable limits.

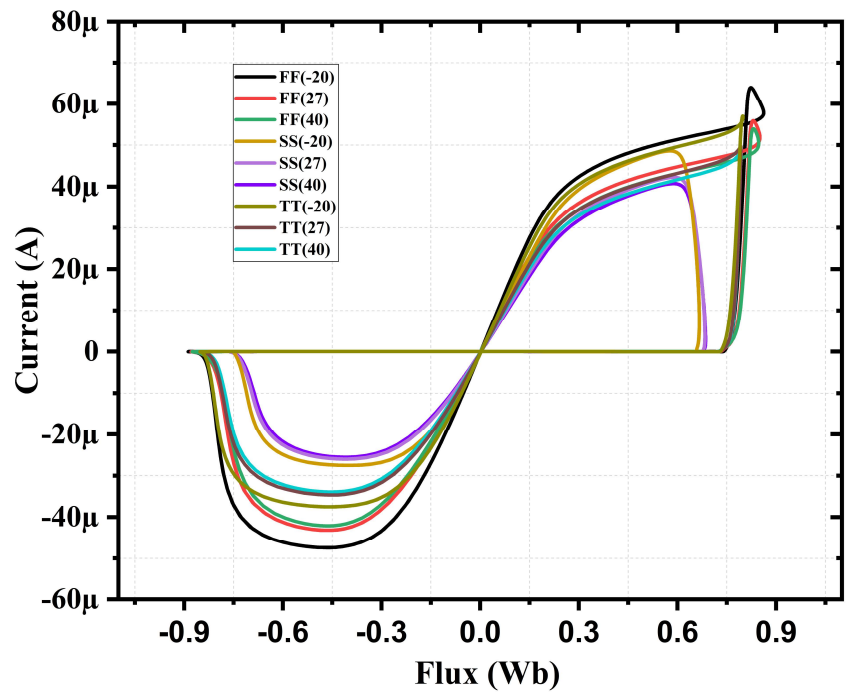

**Fig. 5.11 Meminductor hysteresis loop at various process corner with temperature**

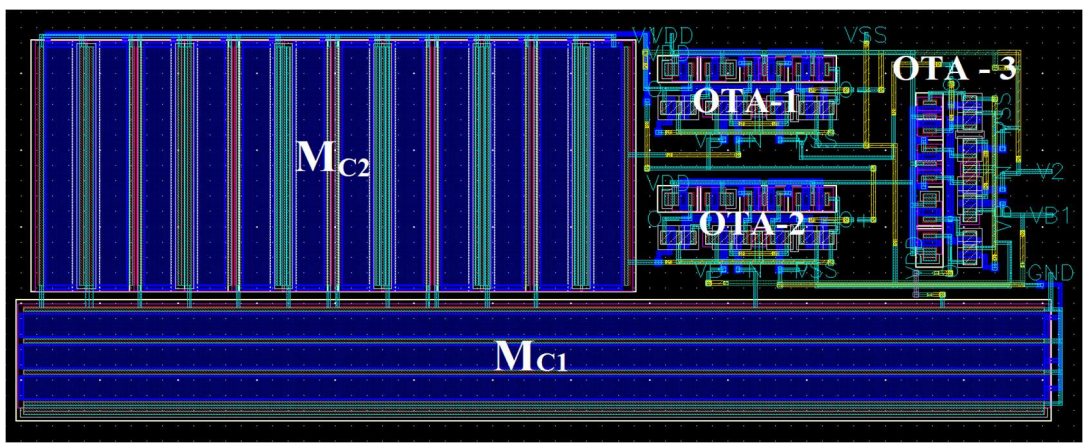

**Fig. 5.12 Layout of the proposed meminductor emulator with 1690.21 µm² of chip area**

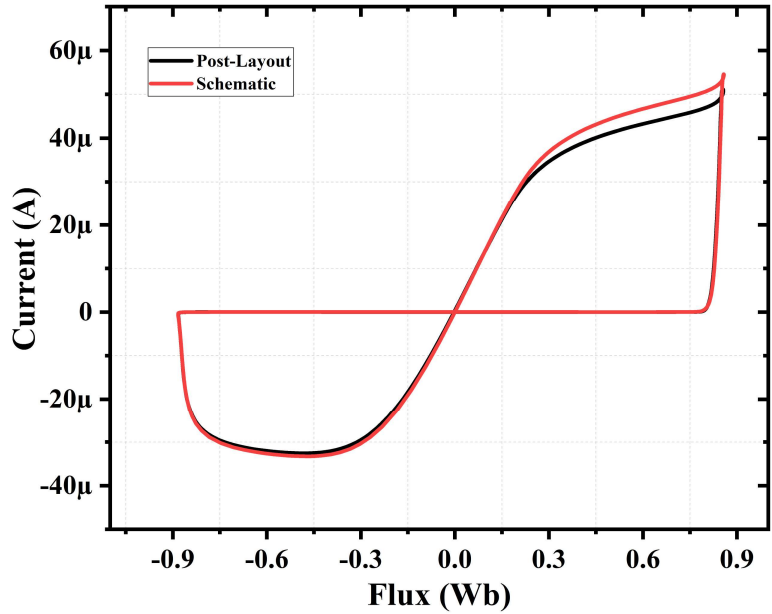

**Fig. 5.13 Pinched hysteresis: schematic versus layout**

### 5.2.5 Experimental Verification

In order to validate the proposed meminductor emulator circuit, physical tests are conducted using readily available components from the open market. The purpose of this section is to conduct experiments on a dual output OTA circuit to verify the accuracy of the proposed circuit topology. As there is no commercially available dual output OTA circuit, the CA3080 integrated circuit is used to implement the dual output OTA and obtain experimental results, as shown in Fig. 5.14 [28]. Since each CA3080 integrated circuit contains two OTA, one integrated circuit is utilized for each dual output OTA element. Additionally, the CA3080 functions as a transconductance amplifier that can be externally controlled by applying a voltage to the bias current terminal. This circuit comprises a three OTA that mimic the behaviour of the proposed meminductor, as shown in Fig. 4.2. It consists of one resistor (Rb), and two capacitors (C1 and C2). It is important to note that the current passing regulates the transconductance gm through the current bias terminal of the CA3080. Fig. 5.15 demonstrates how these components are connected to construct a meminductor on a breadboard for experimental purposes.

The experimental setup consists of a Tektronix AFG3252 signal generator that generates an input signal at various frequencies. The Tektronix TDS2024B digital storage Oscilloscope is employed to study specific signal waveforms, including transients and their accompanying Iin features. The circuit in Fig. 5.15 is powered by a regulated DC voltage from an Rigol DP1308A DC power source.

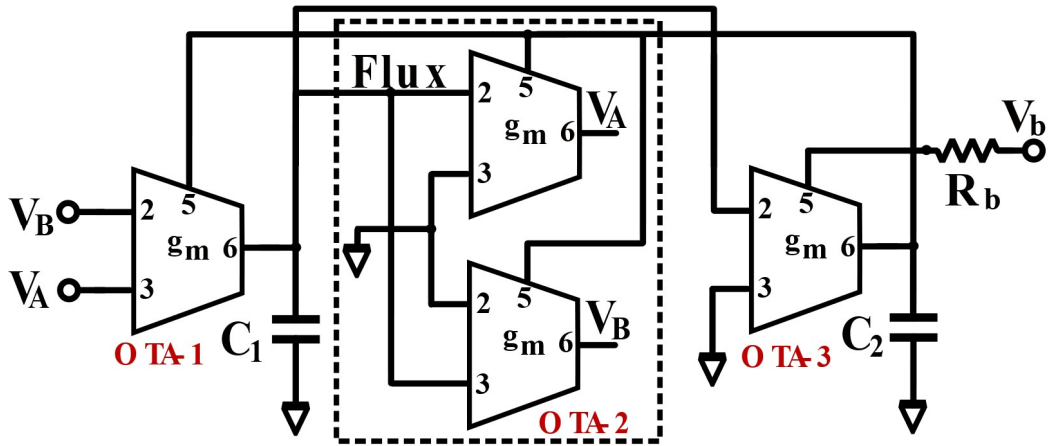

**Fig. 5.14 Meminductor's experimental circuit using the CA3080**

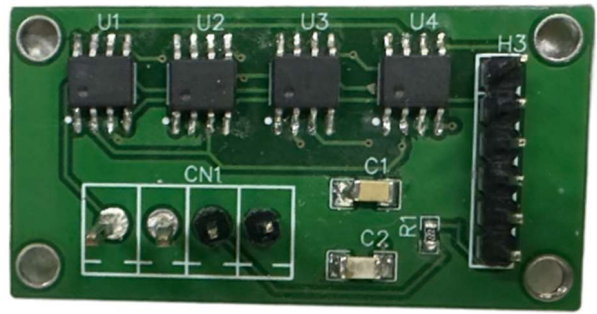

**Fig. 5.15 Proposed meminductor circuit layout on a PCB**

The operating frequencies of the meminductor are adjusted using capacitors and transconductances. In this experiment, the OTA are biased with Vb = 2.2 V using a controlled DC voltage source through Rb = 220 to achieve the desired properties of a meminductor. The process utilizes C1 = 1pF and C2 = 300pF. The CA3080 OTA are supply voltage with VDD = 15 V and VSS = - 15 V. When a 1 MHz signal with a 500 mV peak voltage is applied, the meminductor's transient properties and hysteresis curve are observed, as shown in Fig. 5.16. To obtain the pinched hysteresis curve, one oscilloscope probe was connected to C1 (for flux), while the second probe was connected to pin 6 of CA3080 (OTA 2) to get input current of the proposed meminductor. It is evident that the results of this experiment are consistent with the simulation analysis. The corresponding pinched hysteresis loop for two different frequencies (100 kHz, 6 MHz) is shown in Fig. 5.17. The fundamental characteristics of the meminductor are described by the decrease in the pinched hysteresis lobe area with an increase in frequency. Therefore, the meminductor model is validated by laboratory confirmation, and the results exhibit similar characteristics to those obtained from simulations.

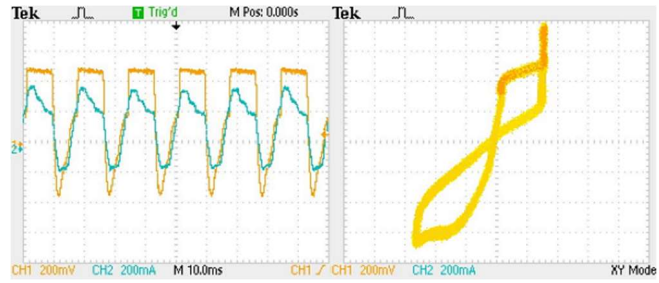

**Fig. 5.16 Floating meminductor (a) Transient response of φ and Iin. (b) Pinched hysteresis loop**

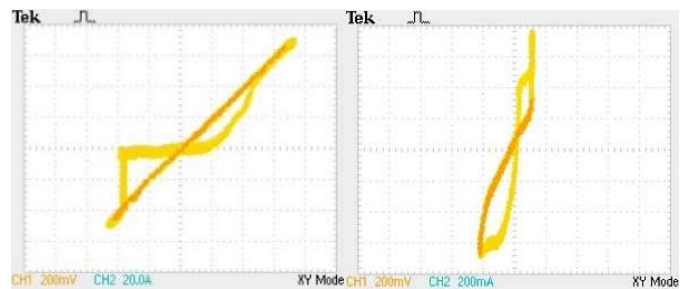

**Fig. 5.17 Hysteresis curve at different frequencies (a) 100 kHz . (b) 6 MHz**

### 5.2.5 Comparison with Prior Works

A brief description of the proposed work with the previously published work is shown in Table 5.4, based on the available physical and performance characteristics:

i. The passive component count is more in [14], [29]-[40], and [36]-[37] which is not feasible because it will consume more power and area consumption is large.

ii. In comparison to [29], [32], [35], [36], [38], [40], and [44]-[45], the overall number of transistors employed in this study is much lower than previous work, which benefits in consuming less space in the chip, and consumes less power consumption.

iii. The proposed emulator is floating and grounded type, whereas reported [29], [31]-[40], [44], and [45], emulators are grounded or floating.

iv. The proposed emulator exhibits better operating frequency compared to [15], and [29]-[45], respectively.

v. In comparison to the work published in [14], [30], [31], [33], [35]-[37], and [45] are verified by only simulation, while the suggested emulator is verified by both experimentally and simulation.

| Ref. | No. of active Components | Transistor Count | Passive component Count | Floating/ Grounded | Technology Used | Operating Frequency | Area | Expt/Sim. |
|---|---|---|---|---|---|---|---|---|
| [29] | CCIIs-4, Analog-1, Multipliers, Opamps-1 | -- | R-6, C-2 | Floating | -- | 960 kHz | 375.4 mm$^2$ | Expt. |
| [30] | AD844- 3, OPAMPs-1, CA3080- 1, Multiplier-1 | >90 | R-8, C-2 | Both | BJT | 22 kHz | -- | Sim. |
| [14] | OTA- 3 | 33 | C-2 | Grounded | 180 nm CMOS | 10 MHz | -- | Sim. |
| [31] | OTA - 3 | 24 | C-2 | Grounded | 180 nm CMOS | 100 kHz | -- | Sim. |
| [32] | OTA-2, Multiplier-1 | 68 | R-2, C-2 | Grounded | PSPICE | 10 kHz | -- | Both |
| [33] | VDTA – 1, OTA - 1 | 27 | C - 2 | Grounded | 180 nm CMOS | 500 kHz | -- | Sim. |
| [34] | VDTA – 1, OTA - 1 | 27 | C - 2 | Grounded | 180 nm CMOS | 20 MHz | 2350.67 μm$^2$ | Both |
| [35] | VDCC - 1 | 44 | M$_R$-1, C-1 | Floating | 180 nm CMOS | 700 kHz | -- | Sim. |
| [36] | CCIIs – 3, Adder-1, Multiplier-1 | >108 | R-3, C-2 | Grounded | -- | 10 Hz | -- | Sim. |
| [37] | DOCCII-2 | -- | R-1, C-1, M$_R$-1 | Floating | BJT | -- | -- | Sim. |
| [38] | DOCCII- 1, CCII- 1, Multiplier-1 | >40 | R-2, C- 2, L - 1 | Grounded | 180 nm CMOS | 300 Hz – 700 Hz | -- | Both |
| [39] | VDTAs-2 | 32 | C-2 | Floating | 180 nm CMOS | 1.5 MHz | -- | Both |
| [40] | CCIIs-2, OTA-1 | 43 | R-2, C-2 | Grounded | 180 nm TSMC | 900 kHz | 8061 μm$^2$ | Both |
| [41] | MVDCC-1, OTA-1 | 39 | R-1, C-2 | Floating | CMOS | 300 kHz | -- | Both |
| [42] | DDCCs-2, Multiplier-1 | 100 | R-2, C-1 | Grounded | 0.35 μm TSMC | Few kHz | -- | Sim. |
| [43] | CCIIs-2, OTA-2 | 43 | R-1, C-2 | Both | 180 nm CMOS | 10 MHz | -- | Both |
| [44] | VDTA-1, DOCCII-1 | 37 | R-1, C-2 | Floating | PSPICE | 100 kHz | -- | Both |
| [45] | VDCC-2 | 50 | C-2 | Both | 180 nm TSMC | 50 MHz | >2441 μm$^2$ | Both |
| **Proposed work** | **OTA-3, MOS-CAP-2** | **37** | **0** | **Both** | **180 nm TSMC** | **60 MHz** | **1690.21 μm$^2$** | **Both** |

CCII-Second generation current conveyor, OPAMP- Operational amplifier, VDTA- Voltage differencing trans-conductance amplifier, OTA- Operational transconductance amplifier, DOCCII- Double output second generation current conveyor, VDCC- Voltage Differencing Current Conveyor, MVDCC- Modified Voltage Differencing Current Conveyor, DDCC- Differential Difference Current Conveyor, Expt- Experiment, Sim - Simulation

**Table 5.4 Comparison between the proposed meminductor emulator with prior works.**

vi. It also showcases approx. less area than [29], [34], [40] and [45], which is its nearest contemporary meminductor in terms of area and operating frequency, whereas it is more power efficient than [29] and [40].

It should be noted that many meminductor emulators [14], [29], [31]-[40], and [44] only provide SPICE or grounded meminductor simulations, use more power, consume large area, have low operating frequency, and don't offer real-world meminductor applications. Finally, there is a dearth of literature on meminductor design, which expands meminductor design's options.

# Chapter 6

# Conclusions

## 6.1 Summary

This paper presents a comprehensive Python framework for evaluating large-scale deep neural networks (DNN) on memristive and memcapacitive crossbar systems, addressing various non-idealities. The framework incorporates device-level considerations, including conductance, capacitance cycle-to-cycle, and device-to-device variations. Testing involved an 8-layer VGG network on a 128 × 128 RRAM array, resulting in memristive and memcapacitive vector-matrix multiplication (VMM) accelerators achieving high training accuracies of 90.02% and 91.03% on the CIFAR-10 dataset, respectively. The impact of factors like line resistance and variations on training accuracy was observed and effectively managed. Detailed hardware estimation for TiOx-based memristor and Si-based memcapacitor VMM accelerators is provided.

Additionally, this paper proposes a meminductor model using three OTAs and two MOS capacitors. The model is validated through extensive testing in diverse scenarios using Cadence Virtuoso with TSMC 180 nm PDK. Experimental validation, employing readily accessible components, aligns with theoretical and numerical analyses, validating the emulator's accuracy. The meminductor emulator exhibits a simpler circuit design, lower energy consumption, and higher operating frequencies (up to 60 MHz, consuming 0.337 mW), leveraging only MOS transistors for easy silicon fabrication.

## 6.2 Future Work

The current research lays a solid foundation for future investigations in mem-elements applied to neural network hardware. Firstly, delving into more sophisticated neural network architectures beyond the simplified 8-layer VGG network promises insights into the scalability and adaptability of memristive and memcapacitive crossbar systems. Optimization of the framework to handle non-idealities more effectively in these devices, such as refining strategies for managing factors like line resistance, cycle-to-cycle, and device-to-device variations, stands as a crucial next step. Extending the application of memristive and memcapacitive crossbar systems to real-world scenarios and diverse datasets will further validate the generalizability and robustness of the proposed frameworks. Exploring physical hardware implementations of the meminductor emulator and scaling its operation across different technology nodes will enhance the understanding of its real-world applicability.

Additionally, investigating the integration of memristive and memcapacitive crossbar systems into edge computing devices and further benchmarking exercises against state-of-the-art hardware implementations will contribute to positioning these technologies within the broader landscape of neural network accelerators. Through these avenues, future research can continue to propel the field forward, addressing practical challenges and advancing the efficiency of computing systems.

# Acknowledgements

I extend my deepest gratitude to my esteemed supervisor, Prof. Byun-Geun Lee, whose unwavering guidance, continuous support, and remarkable patience have been instrumental throughout my pursuit of a Master's degree. Prof. Lee's profound knowledge and extensive experience not only enriched my academic endeavors but also significantly influenced my daily life. His dedication to fostering the research pursuits of his students has been a constant source of motivation, ultimately leading me to this significant milestone. I am indebted to the members of my thesis committee, Prof. Byung-Geun Lee, Prof. Moongu Jeon, and Prof. Chang-Soo Park, for their invaluable contributions. Their insightful suggestions and rigorous examination of my thesis have played a pivotal role in refining and enhancing the quality of my research work. Their collective wisdom has left an indelible mark on my academic journey. A special note of appreciation goes to my colleagues at the Analog and Mixed Signal Integrated Circuit Laboratory. Their unwavering support and collaborative efforts have fostered a conducive research environment, enriching my overall academic experience.

I reserve my deepest appreciation for my family, whose enduring love, unwavering support, and blessings have been my pillar of strength. To my parents, Mr. Suresh Kumar Singh, and Mrs. Kamlawati Singh, to my sister Nidhi Singh and her husband, Dilip Singh, and my brother Abhimanyu Singh. I extend heartfelt thanks for their constant support and encouragement throughout my years of study. Their belief in my capabilities has been a driving force behind my academic achievements and the completion of this thesis. Lastly, I express my gratitude to the Almighty for guiding me through this transformative journey. This thesis marks not only an academic accomplishment but a testament to the collective efforts of those who have played an integral role in shaping my academic and personal growth.

# ANKUR SINGH

123 Cheomdangwagi-ro, GIST, Gwangju, Republic of Korea - 61005.
Email: ankur1660@gmail.com , ankursingh@gm.gist.ac.kr
Google Scholar

**OBJECTIVE:**

A passionate student interested in Integrated circuits, In-memory Computing and Neuromorphic circuits.

**EDUCATION:**

M.S in Electrical Engineering and Computer Science, Gwangju Institute of Science and Technology, Gwangju, South Korea.   **3.71/4.5, 91% 2022 – 2024**

B. Tech in Electronics and Communication Engineering, Indian Institute of Information Technology, Guwahati, India.   **8.37/10, 83.7% 2017 – 2021**

**RELEVANT COURSES:**

Mixed Signal Integrated Circuit Design, IC Design, VLSI design, System on Chip (SoC), Microwave Engineering, Analog Integrated Circuits, Analog Circuits, Embedded Systems, Digital Design, Complex Analysis, Semiconductor and Devices, Digital Communication, Control Systems, Linear Algebra, Calculus and Differential Equation

**RELEVANT PROJECTS AND INTERNSHIP:**

- **Research Assistant** - Analog and Mixed Signal Integrated Circuits Lab, Gwangju Institute of Science and Technology, South Korea – Working on neuromorphic circuit and In-memory Computing.   **Mar 2022 - Present**
- **Summer Research Intern** - Khalifa University, Abu Dhabi, UAE – Worked independently on, current to voltage integrator using a Trans-impedance amplifier in Cadence.   **July 2020 - Oct 2020**
- **Summer Research Intern** - Indian Institute of Technology, Jodhpur: Worked Independently on implementation of a Temperature sensing circuit-using CMOS in Cadence Virtuoso and performed pre-simulation and post layout simulation.   **May 2019 - July 2019**
- **Summer Research Intern** - CETPA InfoTech Pvt. Ltd., Noida: During this Internship, I worked independently on a smart attendance system using RFID and GSM.   **May 2018 - July 2018**

**PAPERS/PUBLICATIONS:**

- **A. Singh**, and B.G. Lee "Electronically Tunable High-Frequency Meminductor Emulator for Neuromorphic Systems" IEEE Transactions on CAS I.   [**Revision**]
- A. M. Z. Khaki, S. Yang, H. Kim, **A. Singh**, and B. G. Lee "A Resource-Efficient Multi-Function Embedded Eye Tracker System Implemented on FPGA" IEEE Access, 2023.
- **A. Singh**, and B. G. Lee "Framework for In-memory Computing based on Memristor and Memcapacitor for On-Chip Training" IEEE Access, 2023.
- M. Ghosh, **A. Singh**, S. S. Borah, J. Vista, A. Ranjan and S. Kumar "MOSFET-Based Memristor for High-Frequency Signal Processing" IEEE Transactions on Electron Devices, 2022.
- S. S. Borah, **A. Singh**, M. Ghosh, and A. Ranjan "Electronically Tunable Higher-Order Quadrature Oscillator Employing CDBA" Microelectronics Journal, 2020.
- M. Ghosh, S. S. Borah, **A. Singh**, and A. Ranjan "Third Order Quadrature Oscillator and its Application using CDBA" Analog Integrated Circuits & Signal Processing, 2020.
- **A. Singh**, S. S. Borah, and M. Ghosh "Simple Grounded Meminductor Emulator using Transconductance Amplifier," 2021 IEEE International Midwest Symposium on Circuits and Systems (MWSCAS), USA.
- S. S. Borah, **A. Singh**, M. Ghosh "A Novel Low-Power Electronically Tunable Higher-Order Quadrature Oscillator using CDBA," 34th International Conference on VLSI Design.

- **A. Singh**, S. S. Borah, and M. Ghosh "Current Differencing Buffered Amplifier Based Memristive Quadrature Oscillator," 20th International Conference on Electronics, Information, and Communication (ICEIC 2021).
- **A. Singh**, S. S. Borah and M. Ghosh "Single VDTA Based Grounded Memristor Model and Its Applications" 2020 IEEE REGION 10 CONFERENCE (TENCON).
- S. S. Borah, **A. Singh**, and M. Ghosh "CMOS CDBA Based 6th Order Inverse Filter Realization for Low-Power Applications" 2020 IEEE REGION 10 CONFERENCE (TENCON).
- **A. Singh**, S. S. Borah and M. Ghosh, "CDBA based Quadrature Sinusoidal Oscillator with Non-Interactive Control," 2020 International Conference on Communication and Signal Processing (ICCSP), Chennai, India, 2020, pp. 0033-0037.
- S. S. Borah, **A. Singh**, M. Ghosh (2020) "Two Novel Configurations of Electronically Tunable Quadrature Sinewave Oscillator Using CDBA," Electronic Systems and Intelligent Computing. Lecture Notes in Electrical Engineering, vol 686. Springer, Singapore.

## ACHIEVEMENTS AND ACTIVITIES:

- Received scholarship from the South Korea Government (GKS) to pursue master's degree at Gwangju Institute of Science and Technology, South Korea       **March 2022 - February 2024**
- Reviewer at IEEE Access       **June 2023 - Present**
- Worked as a volunteer at Universal Culture Center where my role entailed actively advocating for the rights of immigrants in South Korea.       **June - July 2023**
- Was part of the MIT COVID-19 hackathon: Beat the Pandemic II, where me and my team of 3 worked on a remote cleaning and high efficiency robot that used ultraviolet light (UVC), for prevention from any virus.       **May 2020**
- Was part of, in India, online hackathon, where me and my team of 3 were, among top 300 in India for designing a low-cost non-contact thermal scanner.       **April 2020**
- Served as the Department Undergraduate Program Committee's (DUPC) student member for the academic year. IIITG.       **November 2018 - November 2019**
- Served as the General Secretary of Sports manage and organize all the sports in the academic year. IIITG.       **September 2018 - September 2019**
- Worked as a coordinator of the Robotic Club. IIITG.       **September 2018 - September 2019**
- Worked as a coordinator of Football Sports. IIITG.       **September 2017 - September 2018**

## REFERENCES:

1. **Dr. Byung-Geun Lee (Professor)**
   Contact: +82-62-715-3231
   School of Electrical Engineering and Computer Science (EECS), GIST, Gwangju, 61005, Republic of Korea
   Email: bglee@gist.ac.kr

2. **Dr. Sung-Min Hong (Associate Professor)**
   School of Electrical Engineering and Computer Science (EECS), GIST, Gwangju, 61005, Republic of Korea
   Email: smhong@gist.ac.kr

3. **Dr. Shekhar Suman Borah (Research Associate)**
   Bhabha Atomic Research Center, Mumbai, 400085, India
   Email: ssborah@barc.gov.in

4. **Dr. John Vista (Post-Doctoral Fellow)**
   Indian Institute of Technology Bombay, Powai, Mumbai, 400078, India
   Email: 20003147@iitb.ac.in